\documentclass[review]{elsarticle}
\usepackage{caption}
\usepackage{multirow}
\usepackage{subcaption}
\usepackage{graphicx}
\usepackage{wrapfig}
\usepackage{epstopdf}
\usepackage{array}
\usepackage{amsmath}
\usepackage{tikz}
\usepackage{setspace}
\usepackage[symbol]{footmisc}
\usepackage[ruled,vlined]{algorithm2e}
\usepackage{lineno,hyperref}
\usepackage[margin=3cm]{geometry}
\usepackage{xcolor}

\begin{document}

\begin{frontmatter}

       






\title{Rice Diseases Detection and Classification Using Attention Based Neural Network and Bayesian Optimization}

\author{Yibin Wang$^{a}$, Haifeng Wang$^{a,}$\footnote{Corresponding author. Email: wang@ise.msstate.edu}, Zhaohua Peng$^{b}$\\
\normalsize \normalfont	 $a$ Department of Industrial and Systems Engineering \\ Mississippi State University, Mississippi State, MS 39762\\ $b$ Department of Biochemistry, Molecular Biology, Entomology and Plant Pathology\\ Mississippi State University, Mississippi State, MS 39762\\
}





\begin{abstract}
In this research, an attention-based depthwise separable neural network with Bayesian optimization (ADSNN-BO) is proposed to detect and classify rice disease from rice leaf images. Rice diseases frequently result in 20 to 40 \% corp production loss in yield and is highly related to the global economy. Rapid disease identification is critical to plan treatment promptly and reduce the corp losses. Rice disease diagnosis is still mainly performed manually. To achieve AI assisted rapid and accurate disease detection, we proposed the ADSNN-BO model based on MobileNet structure and augmented attention mechanism. Moreover, Bayesian optimization method is applied to tune hyper-parameters of the model. Cross-validated classification experiments are conducted based on a public rice disease dataset with four categories in total. The experimental results demonstrate that our mobile compatible ADSNN-BO model achieves a test accuracy of 94.65\%, which outperforms all of the state-of-the-art models tested. To check the interpretability of our proposed model, feature analysis including activation map and filters visualization approach are also conducted. Results show that our proposed attention-based mechanism can more effectively guide the ADSNN-BO model to learn informative features. The outcome of this research will promote the implementation of artificial intelligence for fast plant disease diagnosis and control in the agricultural field.

\end{abstract}

\end{frontmatter}

\section{Introduction}

Disease is a leading factor that affects the growth and development of rice. Rice diseases are causing a large quantity of loss in grain yield. FAO (Food and Agriculture Organization, USA) estimates that plant diseases cost the global economy around \$220 billion annually and between 20 to 40 percent of global crop production is lost to diseases and pests \citep{agrios2005plant}. Identifying the plant diseases rapidly and precisely is critical to reduce the economic losses. However, disease diagnosis and treatment usually require special skills and equipment, which farmers often don’t have, thus resulting in a delay of treatment while waiting for experts to diagnose and treat the disease or due to misdiagnosis. The rapid advancement in mobile device and digital camera technologies have made it highly convenient for people to obtain digital images of crop plants and organs with diseases.  Meanwhile, machine learning, deep learning, and computer vision technologies have made image processing and disease diagnosis become feasible as shown in many studies \citep{wang2019deep, lu2018dual, zhang2018medical}. 
Pesticide treatments are used as the main methods to control corp diseases. However, the use of pesticides highly depends on diagnosis results, and unrestrained pesticides usage causes considerable environmental damage as well \citep{wu2017characterization}. Therefore, precise and timely diagnosis of the diseases is critical. Currently, the diagnosis of rice diseases is often performed manually, and the process can be extremely labor intense and time-consuming. To overcome the drawbacks, computer-aided diagnosis is applied as an important tool in detecting and classifying rice diseases.

The convolutional neural network (CNN) is a type of popular deep neural networks (DNNs) that has demonstrated high generalization performance in many image analysis studies. For instance, an attention based deep residual network is proposed to detect the type of infection in tomato leaves \citep{Tomato}. Performance of different deep learning based approaches is studied to classify skin diseases from colored digital photographs \citep{goceri2019skin, goceri4}. Neural network modeling associated with adaptive neuro-fuzzy inference system (ANFIS) is used to describe and predict performance of a horizontal ground-coupled heat pump (GCHP) system \citep{esen2008performance, esen2009artificial, esen2017modelling}. To obtain high performances from deep networks, appropriate loss functions, activation functions and batch sizes should be used \citep{Goceri, Goceri2}. Challenges and recent solutions for image segmentation in deep learning are well-explained \citep{Goceri3}. Even though DNN techniques have been developed rapidly, seldom has been implemented for rice disease detection. In other words, a proper model designated for rice disease classification is lacking. In this study, we focus on analyzing the extracted rice disease pattern from a deep learning model. An optimization-based attention mechanism is developed to help the deep neural networks pay more attention to the critical features. In addition, our model is designed for running on portable devices, such as mobile phones and single board computers, which is a significant improvement from other studies. To extract diagnostic features, a large amount of rice leaf images are typically required in training deep learning models. In this research, we propose an attention mechanism to enhance the DNN model learning process to guide the model focus more on extracting more informative features.

Visualizing and understanding how the neural networks learning are quite essential. In practice, there are many helpful approaches, for instance, intermediate activation visualization, convolution filter visualization, heat map and saliency map. The aim of the feature visualization is to reveal how the model represents the visual world from its perspective. Additionally, it helps us to understand how a pre-trained convolutional network extracts feature, and show intuitively desirable properties such as compositionality, increasing invariance and class discrimination as the layers being ascended. These visualization techniques can also be used to debug problems with the model to obtain better results \citep{zeiler2014visualizing}. 

The contributions of our work presented in this article are listed as follows: (1) a novel CNN model structure applying attention-based depthwise separable convolution is proposed and tested; (2) the proposed ADSNN-BO model outperforms all the tested state-of-the-art models in classification accuracy and other performance measurements with regard to rice disease classification mission; (3) the performance of ADSNN-BO model is analyzed and compared with other deep learning models using filter visualization approach. The structure of this paper is organized as following. Section 2 provides an overview of deep learning models and implementations in rice diseases. Section 3 describes different learning architectures as well as modifications for rice diseases classification. Experimental results and related performance evaluation of the deep learning models are illustrated in Section 4. Approaches used to extract samples and other image processing steps are also discussed in Section 4. The research findings and future work are concluded in Section 5.

\section{Literature Review}

Multiple deep learning techniques have been developed and implemented with rice diseases detection over the past few years.


\begin{table}[!htp]
\centering
\caption{Publications related to rice diseases detection by machine learning and deep learning}
\label{table 1}
\resizebox{\textwidth}{!}{%
\begin{tabular}{ccccccc} 
\hline
Year & Number of Diseases & Type of Disease                                                                                                         & Methods                 & Highest Accuracy ($\%$) & \multicolumn{1}{l}{ Filter Visualization} & Reference \\ \hline
2016  & 1                  & Bakanae disease                                                                                                         & SVM, GA                 & 87.90                & $\times$                                                  & \citet{chung2016detecting}  \\
2017  & 10                 & \begin{tabular}[c]{@{}c@{}}Rice diseases including rice blast, rice false smut, \\ rice brown spot, etc.\end{tabular}   & CNN
& 95.48                &                                                & \citet{lu2017identification}   \\
2018  & 9                  & \begin{tabular}[c]{@{}c@{}}Rice diseases including false smut, \\ brown plant hopper, etc.\end{tabular}                 & CNN              & 93.30                &                                              & \citet{rafeed2018identification}   \\
2019  & 1                  & Rice blast                                                                                                              & CNN, SVM, LBPH          & 95.83                & $\times$                                                & \citet{liang2019rice}   \\
2019  & 4                  & \begin{tabular}[c]{@{}c@{}}Rice diseases including rice blast, bacterial blight, \\ sheath rot, brown spot\end{tabular} & DNN$\_$JOA, ANN, DAE      & 97.00           & $\times$                                                  & \citet{ramesh2020recognition}   \\
2019  & 1                  & Rice blast                                                                                                              & RiceTalk                & 89.40                & $\times$                                                  & \citet{chen2019ricetalk}   \\
2019  & 1                  & Different levels of contamination of grain discoloration                                                                & InceptionV3             & 88.20                & $\times$                                                  & \citet{duong2019classification}   \\
2019  & 3                  & Rice blast, bacterial leaf blight, sheath blight                                                                        & AlexNet, CNN, SVM       & 91.37               & $\times$                                                  & \citet{shrivastava2019rice}  \\ \hline
\end{tabular}%
}
\end{table}

Liang et al. reported a novel rice diseases identification method based on deep CNN techniques \citep{lu2017identification}. They used a dataset of 500 natural images of diseased and healthy rice leaves and stems captured from rice experimental field with 10 classes of common rice diseases. Under the 10-fold cross-validation strategy, the proposed CNNs-based model achieves an accuracy of 95.48\%. Rahman et al. applied state-of-the-art large scale architectures, VGG16 and InceptionV3, and fine tuned for detecting and recognizing rice diseases and pests \citep{rafeed2018identification}. Their experimental results showed the effectiveness of these models with real datasets. Additionally, since large scale architectures were not suitable for mobile devices, a two-stage small CNN architecture was proposed and compared with the state-of-the-art memory efficient architectures MobileNet, NasNet Mobile and SqueezeNet. They achieved the desired accuracy of 93.3\% with a significantly reduced model size. Liang et al. presented a rice blast feature extraction and disease classification method based on deep CNN \citep{liang2019rice}. They also established a rice blast disease dataset with the assistance of plant protection experts, which can be combined with other rice disease images to build a content-rich dataset. They conducted comparative experiment and applied t-SNE approach and found that the high-level features extracted by CNN were more discriminative and representative than local binary patterns histograms (LBPH) and Haar-WT (wavelet transform). Their quantitative analysis results indicated that CNN and CNN with support vector machine (SVM) had almost the same performance, which was better than that of LBP + SVM and Haar-WT + SVM.

Ramesh and Vydeki utilized the image processing techniques to minimize the reliance on the farmers to protect the agricultural products \citep{ramesh2020recognition}. They proposed a recognition and classification of paddy leaf diseases using optimized DNN with Jaya Algorithm. Their images of rice plant leaves were directly captured from the farm field for healthy, bacterial blight, brown spot, sheath rot and blast diseases. In preprocessing, for the background removal the RGB images were converted into HSV images and based on the hue and saturation parts binary images were extracted to split the diseased and non-diseased part. Classification of diseases is carried out by using Optimized Deep Neural Network with Jaya Optimization Algorithm (DNN\_JOA). Their experimental results were evaluated and compared with ANN, DAE and DNN. They achieved high accuracy of 98.9\% for the blast affected, 95.78\% for the bacterial blight, 92\% for the sheath rot, 94\% for the brown spot and 90.57\% for the healthy leaf image. Chen et al.'s work studied rice blast detection and classification by using the Internet of Things (IoT) and artificial intelligence (AI) \citep{chen2019ricetalk}. Their research was developed to overcome the gap that existing AI and IoT studies detecting plant diseases by images or nonimage hyperspectral data requires manual operations to obtain the photographs or data for analysis. Besides, image detection typically is too late as rice blast may already spread to other plants. They explored the RiceTalk project that utilized nonimage IoT devices to detect rice blast based on an IoT platform for soil cultivation. Unlike the image-based plant disease detection approaches, their agriculture sensors generated nonimage data that can be automatically trained and analyzed by the AI mechanism in real time. Their approach reduced the platform management cost significantly to provide real-time training and predictions. In their implementation, the accuracy of the RiceTalk prediction on rice blast is 89.4\%.

Since machine learning techniques are commonly applied before deep learning becomes popular. Rehman et al. reviewed statistical machine learning technologies with machine vision systems in agriculture \citep{rehman2019current}. Two types of statistical machine learning techniques, supervised and unsupervised learning, have been concentrated on. Their paper comprehensively surveyed current application of statistical machine learning techniques in machine vision systems, such as Naive Bayes (NB), discriminant analysis (DA), k-nearest neighbour (kNN), support vector machine (SVM), etc. They analyzed each technique potential for specific application and represents an overview of instructive examples in different agricultural areas. Suggestions of specific statistical machine learning technique for specific purpose and limitations of each technique were also given. Duong-Trung et al. implemented the idea of transfer learning to classify grain discoloration disease of rice \citep{duong2019classification}. The discoloration was considered as a potential risk to the rice-producing countries. Transfer learning is a method of learning in a new prediction task with the transferred knowledge from a related task that has already been learned. By utilizing InceptionV3 model of CNNs and transferred weights from ImageNet to train their collected data, the classification accuracy of 88.2\% was achieved. Shrivastava et al. employed transfer learning of deep CNN to classify rice plant diseases \citep{shrivastava2019rice}. Their proposed model was able to classify rice diseases with classification accuracy of 91.37\% for 80\%- 20\% training-testing partition. They adjusted the benchmarking of the model making it more appropriate.

Chung et al. proposed an approach to discriminate infected and healthy seedlings nondestructively at the age of 3 weeks using machine vision \citep{chung2016detecting}. Their work concentrated on detecting bakanae disease, which is a seed-borne disease of rice. Infected plants can yield empty panicles or perish. The images of infected and control seedlings were acquired using flatbed scanners to quantify their morphological and color traits. Support vector machine (SVM) classifiers were employed for distinction. A genetic algorithm was also mentioned for selecting essential traits and optimal model parameters for the classifiers. Their proposed approach distinguished infected and healthy seedlings with an accuracy of 87.9\% and a positive predictive value of 91.8\%.

Feature analysis is important in understanding the performance of the deep learning model being used. \citep{lu2017identification} implemented feature maps to visualize the corresponding feature visualization of rice disease image patch. \citep{rafeed2018identification} showed the first convolution activation outputs of the CNN model. The activations retained almost all of the information in the input image, and the intermediate outputs for different classes are also visually different. Since feature analysis is seldom discussed in previous researches, this paper will concentrate more on feature visualization and understanding. More sophisticated deep learning models should rather be understood as an attempt to both search for the minimum necessary ingredients for recognition with CNNs \citep{springenberg2014striving}. The connection between the gradient-based CNN visualisation methods and deconvolutional networks can be established via saliency maps \citep{simonyan2013deep}.

Advanced optimization method is important for improving deep learning model performance. Similarly, the loss function used in a deep network affects performance of the model. Therefore, cross-entropy, Tversky similarity function or combination of them has been applied, such as in \citep{goceri2020deep}, in the literature to solve classification problems. Bayesian Optimization provides an efficient technique based on Bayes Theorem to guide an optimization search problem. The algorithm functions by building a probabilistic model of the objective function which is searched efficiently with an acquisition function. The candidate samples are determined by being evaluated on the objective function \citep{snoek2012practical}.

In summary, a list of related literature has been presented in Table \ref{table 1}. Although deep learning techniques have been used in rice disease diagnosis, studies on models that can easily fit to the design requirements for mobile are still very limited. Also, advanced methods need to be applied to tune hyper-parameters in deep learning models. Insightful model performance evaluation is still missing. Research in analyzing feature characteristics within different models is also not available at the current stage. To search for an optimized mobile device friendly model for rice disease detection, we propose a new model (ADSNN-BO) and compare it with several DNNs, and provide intermediate activations visualization as well as feature patterns visualization of various convolutional layers. Potential methods to evaluate useful convolution filters are also discussed in this paper.

\section{Attention-based Depthwise Separable Neural Network with Bayesian Optimization}

\subsection{Attention-based Depthwise Separable Deep Neural Network}
MobileNet deep learning model embedded with attention augmented layer and Bayesian optimization method is mainly focused on. We here illustrated the new ADSNN-BO model we proposed. Besides, the comparison of other deep learning models, such as VGG16, DenseNet, ResNet, InceptionV3, and Xception model are discussed to classify rice disease images. We have considered inserting attention convolutional layer into original deep learning models. Bayesian optimization is applied when exploring the influence of parameters in a deep learning model. Although different optimization approaches, such as Sobolev gradient, have been used in some recent works \citep{goceri2019diagnosis} to increase generalization performance of deep networks, they cause high computational costs. Therefore, Bayesian optimization has been implemented in this work.

CNN is a multi-layer neural network with a learning architecture which is usually consisted of two parts, including a feature extractor and a trainable classifier. In practice, a CNN learns the values of filters by itself during the training process. More precisely, a convolution layer attempts to learn filters in a three-dimensional space, with two spatial dimensions and a channel dimension. Therefore, a single convolution kernel is conducted with simultaneously mapping cross-channel and spatial correlations.

MobileNet was proposed as an efficient CNN architecture to achieve very small, low latency models that can easily fit to the design requirements for mobile and embedded vision applications \citep{howard2017mobilenets, gocceri2020impact}. The MobileNet model is based on a streamlined architecture using depthwise separable convolutions to build DNNs with light weights. Compared with a standard convolution, a depthwise separable convolution consists of a depthwise convolution and a 1 $\times$ 1 pointwise convolution. The conventional convolution combines inputs into a new set of outputs within one step. The depthwise separable convolution splits this into two layers, a separate layer for filtering and another for combining. To illustrate, a conventional convolutional layer takes $D_{F} \times D_{F} \times M$ as an input and produces a $D_{G} \times D_{G} \times N$ output feature map, where $D_{F}$ is the spatial width and height of an input feature map, M is the number of input channels, $D_{G}$ is the spatial width and height of an output feature map, and N is the number of output channels. The conventional convolutional layer is represented by convolution kernel $K$ of size $D_{K} \times D_{K} \times M \times N$ where $D_{K}$ is the spatial dimension of the kernel.

From the computational cost perspective, standard convolutions have the computational cost of,

\begin{equation} \label{eq1}
\begin{aligned}
D_{K} \cdot D_{K} \cdot M \cdot N \cdot D_{F} \cdot D_{F}
\end{aligned}
\end{equation}

MobileNet architecture addresses each of these terms and their interactions. Depthwise separable convolutions is used to break the interaction between the number of output channels and the size of the kernel. In this way, the model will learn information within one particular channel and interactions between the channels separately. Depthwise separable convolution involves two kinds of layers: 1) depthwise convolutions and 2) pointwise convolutions. The Depthwise convolution refers to a single filter applied for each input channel. Pointwise convolution refers to a simple 1 $\times$ 1 convolution which is used to establish a linear combination of the output of the depthwise layer. Depthwise convolution has a computational cost of,

\begin{equation} \label{eq2}
\begin{aligned}
D_{K} \cdot D_{K} \cdot M \cdot D_{F} \cdot D_{F}
\end{aligned}
\end{equation}

Depthwise convolution performances well inside each particular channel, and pointwise convolution is then employed to address the lack of the relations between channels. Thus, an additional layer that creates a linear combination of the output of depthwise convolution via 1 $\times$ 1 convolution is added to generate new features. Overall, depthwise separable convolution has a computational cost of,

\begin{equation} \label{eq3}
\begin{aligned}
D_{K} \cdot D_{K} \cdot M \cdot D_{F} \cdot D_{F} + M \cdot N \cdot D_{F} \cdot D_{F}
\end{aligned}
\end{equation}

which refers to the summation of the depthwise and pointwise convolutions.

By presenting convolution as a two-step process with filtering and combining. A reduction in computational cost can be achieved as,

\begin{equation} \label{eq4}
\begin{aligned}
\frac{D_{K} \cdot D_{K} \cdot M \cdot D_{F} \cdot D_{F} + M \cdot N \cdot D_{F} \cdot D_{F}}{D_{K} \cdot D_{K} \cdot M \cdot N \cdot D_{F} \cdot D_{F}}
\end{aligned}
\end{equation}

which equals to $\frac{1}{N} + \frac{1}{D_{K}^{2}}$. Therefore, by utilizing depthwise separable convolution module, MobileNet can effectively reduce the model size and computation complexity.

Similar to MobileNet, the methodology behind the Inception module is to make the training process easier and more efficient by clearly sectioning it into a series of operations that independently focuses on cross-channel and spatial correlations. For the InceptionV3 model, Wojna et al. explored some ways of factorizing convolutions in various settings, especially to increase the computational efficiency of the solution \citep{szegedy2016rethinking}. Since Inception networks are fully-convolutional, each weight corresponds to one particular multiplication per activation. Thus, any reduction in computational cost will lead to reduced number of parameters.

Chollet proposed a CNN architecture fully based on depthwise separable convolution layers called Xception. As an improvement regarding to the Inception model, the hypothesis was made that the mapping of cross-channels and spatial correlations in the feature maps can be entirely disengaged \citep{chollet2017xception}. The original Xception model possesses 36 convolutional layers forming the feature extraction base. Compared with the Inception module, it is noted that there are two minor differences. One is the order of the operations. In Xception module, depthwise separable convolutions as usually are implemented to perform first channel-wise spatial convolution and then perform 1x1 convolution, however the Inception model performs the 1x1 convolution first. The other is that the presence or absence of a non-linearity after the first operation. In Inception module, both operations are followed by a ReLU non-linearity, whereas for Xception, depthwise separable convolutions are usually implemented without non-linearities.

\begin{figure}[!h]
\captionsetup[subfigure]{justification=centering}
  \centering
  \includegraphics[width=0.8\textwidth]{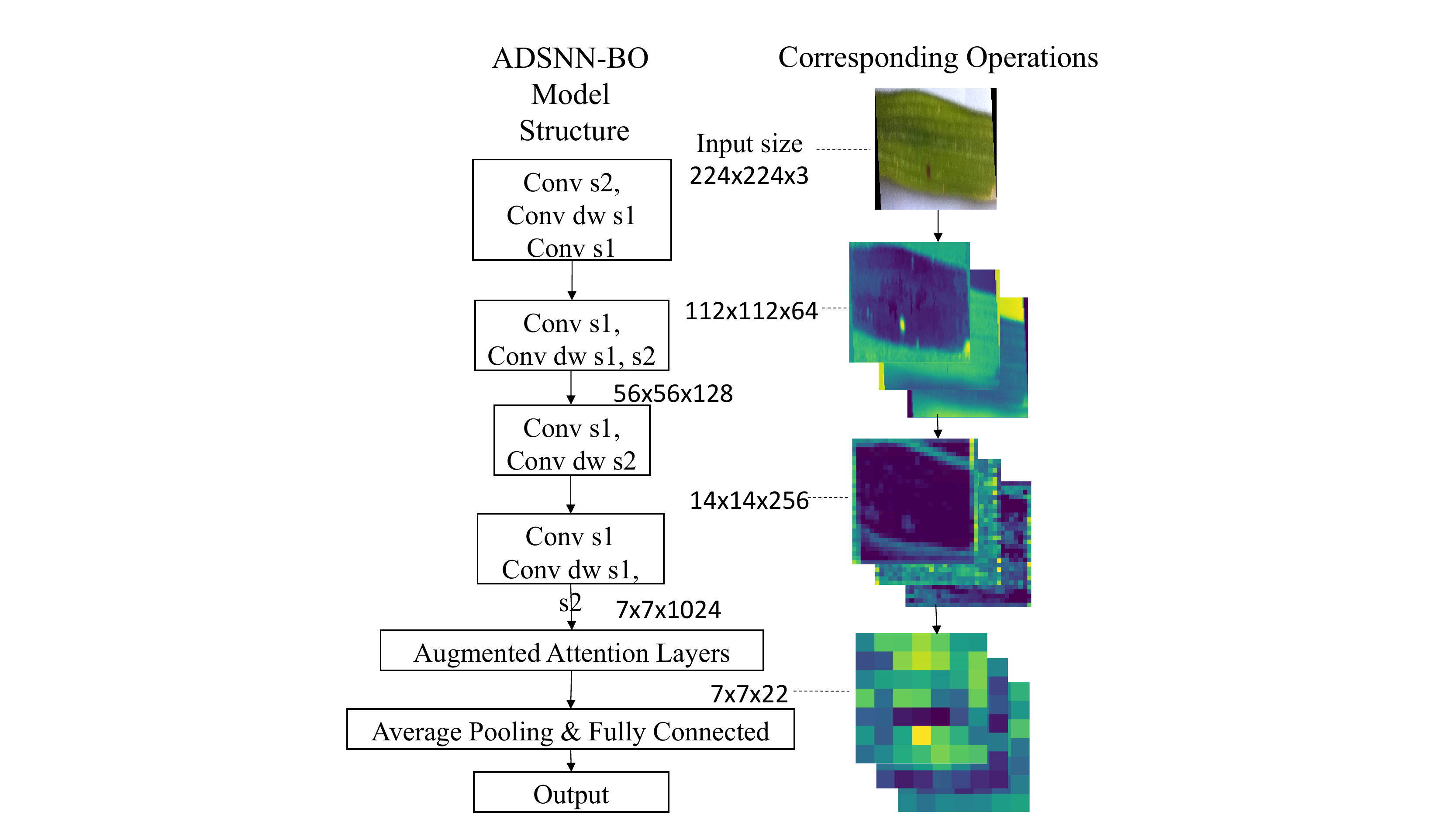}
  \caption{CNN architecture of the proposed model}
  \label{Figure 1}
\end{figure}

To overcome the limitations of trading off accuracy to reduce size and latency, we propose a new model based on depthwise separable neural network embedded with augmented attention mechanism. We take the architecture of MobileNetV1 as the bare-bones of the model, and apply attention layers between the last convolution block and the average pooling layer. We keep the original MobileNet structure as well as the hyperparameters settings. The detailed architecture is shown in Figure \ref{Figure 1}. Each convolution block is specified in the figure. In addition, the outputs of corresponding convolution operations are indicated.

Neural network models are generally referred to as being opaque. This means that they usually fail to explain the reason why a specific decision or prediction was made. CNNs are designed to work with image data, and their structure and function suggest that should be less inscrutable than other types of neural networks. Specifically, the models are comprised of small linear filters and the result of applying filters called activation maps or feature maps. Both filters and feature maps can be visualized. An efficient way to inspect the filters learned by convolutional networks is to display the visual pattern that each filter is meant to respond to \citep{francois2017deep}. This can be conducted with gradient ascent in input space: applying gradient descent to the value of the input image of a convolutional network so that the response of a specific filter is maximized. The procedure usually starts from a blank input image. The resulting image will be one that the chosen filter is maximally responsive to. These filter visualizations reveal a lot about how convolutional layers learn. Each layer in a convolutional network learns a collection of filters so that their inputs can be expressed as a combination of the filters. Further visualization techiques such as heatmap can be followed up to explore and evaluate the model.

\subsection{Augmented Attention Mechanism}

Attention mechanism in deep learning has gathered widespread attention as a computational module for modeling sequences because of its ability to apprehend long distance interactions within a model. Our attention augmented networks do not rely on pre-training of the fully convolutional counterparts and employ self-attention along the entire architecture. Additionally, the architecture is enhanced with the representational power of self-attention over images by extending relative self-attention to two dimensional inputs\citep{bello2019attention}.

For a self-attention procedure, given an input tensor of shape $(H, W, F_{in})$, we flatten it to a matrix $X \in R ^ {HW \times F_{in}}$ and perform multi-head attention. The output of the self-attention mechanism for a single head $h$ can be formulated as:

\begin{equation} \label{eq5}
\begin{aligned}
O_{h}=Softmax(\frac{(XW_{q})(XW_{k})^{T}}{\sqrt{d_{k}^{h}}})(XW_{v})\end{aligned}
\end{equation}

where $W_{q}, W_{k} \in R^{F_{in}\times d_{k}^{h}}$ and $W_{v} \in R^{F_{in}\times d_{v}^{h}}$
are learned linear transformations that map the input $X$ to queries $Q = XW_{q}$, keys $K = XW_{k}$ and values $V = XW_{v}$. $N_{h}$ and $d_{k}$ respectively refer to the number of heads, the depth of values and the depth of queries and keys in multihead-attention. The outputs of all heads are then concatenated and projected again as follows:

\begin{equation} \label{eq6}
\begin{aligned}
M(X)=Concat[O_{1}, ..., O_{Nh}]W^{O}\end{aligned}
\end{equation}

where $W^{O} \in d_{v}\times d_{v}$ is a learned linear transformation. $M(X)$ is then reshaped into a tensor of shape $(H, W, d_{v})$ to match the original spatial dimensions. It is noted that multi-head attention incurs a complexity of $O((HW)^2d_{k})$ and a memory cost of $O((HW)^2N_{h})$ as it requires to store attention maps for each head.

With an original convolution operator with kernel size $k$, $F_{in}$ input filters and $F_{out}$ output filters. The corresponding attention augmented convolution can be expressed as:

\begin{equation} \label{eq7}
\begin{aligned}
AConv(X)=Concat[Conv(X), M(X)]\end{aligned}
\end{equation}

We employed the augmented convolution layers directly after conventional convolutions which are followed by average pooling computation. Similarly to the traditional convolution, the utilized attention augmented convolution is equivalent to translation and can smoothly work on inputs of different spatial dimensions. Effects on attention parameters such as the number of filters and the number of attention layers implemented are explored using Bayesian optimization approach.

\subsection{Bayesian Optimization for Hyperparameter Tuning}

Bayesian optimization works by constructing a posterior distribution of functions which is gaussian process that best describes the function to optimize. As the number of observations grows, the posterior distribution improves, and the algorithm becomes more certain of which regions in parameter space are worth exploring and computing.

\begin{algorithm}[!h]
\SetAlgoLined
 Place a Gaussian process prior on $f$.\\
 Observe $f$ at $n_{0}$ points according to an initial space-filling experimental design.\\
 Set $n = n_{0}$.\\
 \While{$n \leq N$}{
  Update the posterior probability distribution on $f$ using all available data\;
  Let $x_{n}$ be a maximizer of the acquisition function over $x$, where the acquisition function is computed using the current posterior distribution\;
  Observe $y_{n} = f(x_{n})$\;
  Increment $n$\;
  }
  Return a solution: either the point evaluated with the largest $f(x)$, or the point with the largest posterior mean.\
 \caption{Bayesian Optimization Algorithm}
 \label{tab:algorithm}
\end{algorithm}

\begin{figure}[!h]
\captionsetup[subfigure]{justification=centering}
  \centering
  \includegraphics[width=0.95\textwidth]{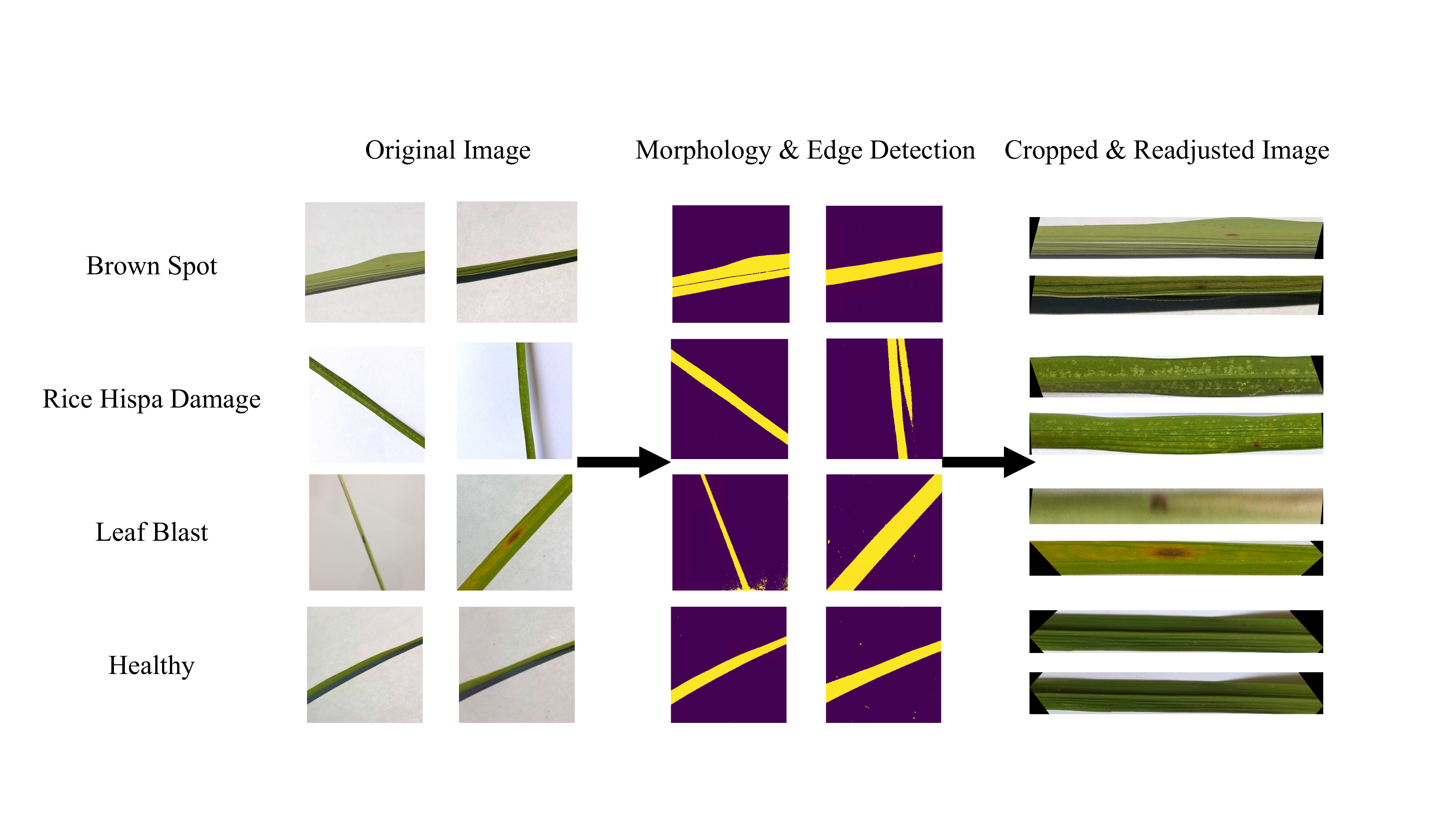}
  \caption{Samples of rice leaf throughout image preprocessing: (a) original samples; (b) images after applying morphology; (c) cropped and position re-adjusted samples}
  \label{Figure 2}
\end{figure}

Bayesian optimization method consists of two major components: one is a Bayesian statistical model to model the objective function, and the other is an acquisition function to decide where to sample next \citep{frazier2018tutorial}. In other words, presenting sampling points in the search space is achieved by acquisition function. The function will trade off exploitation and exploration. Exploitation refers to sample where the Bayesian statistical model predicts a high objective score, and exploration means sampling at locations where the prediction uncertainty is high. Both contribute to high acquisition function values and the aim is to maximize the acquisition function to determine the next sampling point. After evaluating the objective based on an initial space-filling experimental design, often containing points chosen uniformly at random, they are used iteratively to allocate the remainder of a budget of $N$ function evaluations, which is shown in Algorithm \ref{tab:algorithm}. The time complexity for deep learning is $O(w \cdot m \cdot e)$, where $w$ refers to the number of weights, $m$ refers to the number of learning samples, and $e$ refers to the running epochs. For Bayesian optimization, the time complexity is $O(n^{3})$, where n is the number of observations.

In this research, we conduct hyperparameters tuning with regard to the embedded augmented attention mechanism. Particularly, the number of filters that are employed in each attention layer is optimized. For hyperparameter tuning with Bayesian optimization, we construct an objective function that takes the number of filters in each attention layer as input and returns a test accuracy score. Throughout the optimization procedure, the best combination of the filter number for each attention layer can be determined.

\begin{table}[!h]
\centering
\caption{Detailed category information of the rice disease dataset}
\label{tab:dataset}
\begin{tabular}{cc}
\hline
Category          & Number of samples \\ \hline
Brown spot        & 523               \\
Rice hispa damage & 565               \\
Leaf blast        & 779               \\
Healthy           & 503               \\
Total             & 2370              \\ \hline
\end{tabular}
\end{table}

\section{Experiments}

\subsection{Data Preparation}

In this paper, we focus on three specific types of rice diseases, which are brown spot, rice hispa damage, and rice leaf blast. The typical manual diagnosis process is based on visualization symptom of each disease. Brown spots are round to oval, dark-brown lesions with yellow halo. As lesions enlarge, they remain round, with center area necrotic, gray and the lesion margin reddish-brown to dark brown. Rice hispa damage scrapes the upper surface of leaf blades leaving only the lower epidermis. The disease also tunnels through the leaf tissues. When damage is severe, plants become less vigorous. The rice hispa damage can be directly detected by visualizing the bug on the leaf. For rice leaf blast, lesions varying from small round, dark spots to oval spots with narrow reddish-brown margins and gray or white center. Spots become elongated, diamond-shaped or linear with wit pointed ends and gray dead areas in the center surrounded by narrow reddish-brown. We conducted our study on a dataset with 2370 rice leaf samples in total, including all the three classes as well as the healthy rice samples \citep{HuyDo2019}. The detailed class information is provided in Table \ref{tab:dataset}. 100 samples are selected in each category for training and testing. Additional image preprocessing methods including morphology, edge detection and cropping are involved in sample preprocessing.

\begin{table}[!h]
\centering
\caption{The performance comparison of different deep learning models}
\label{tab:performance}
\resizebox{\textwidth}{!}{%
\begin{tabular}{cccccc}
\hline
Model                   & Precision (SD) (\%)  & Recall (SD) (\%)     & F-1 Score (SD) (\%)  & Test Accuracy (SD) (\%) & Training Time (SD) (min) \\ \hline
VGG16                   & 22.3 (17.08)        & 62.5 (26.30)          & 35.5 (2.38)          & 67.12 (5.84)            & \textbf{2.161 (0.03)}   \\
ResNet50                & 62.8 (25.24)         & 76.8 (14.04)         & 64.0 (12.98)         & 79.32 (6.12)            & 5.164 (1.23)            \\
DenseNet121             & 84.0 (12.67)         & 76.0 (16.31)         & 78.2 (7.98)          & 88.27 (4.77)            & 9.354 (0.48)            \\
MobileNet               & 85.6 (12.95)         & \textbf{90.2 (9.28)} & 86.8 (6.22)          & 91.36 (2.82)            & 3.325 (0.41)             \\
Inception V3            & 70.8 (18.58)         & 64.8 (18.19)         & 65.2 (13.77)         & 83.20 (5.08)            & 5.616 (0.38)             \\
Xception                & 88.2 (9.68)          & 80.4 (11.06)         & 84.2 (10.76)         & 90.38 (2.74)            & 7.968 (0.80)             \\
\textbf{ADSNN-BO} & \textbf{92.6 (8.17)} & 87.4 (8.26)          & \textbf{89.6 (4.62)} & \textbf{94.65 (2.07)}   & 3.368 (0.50)            \\ \hline
\end{tabular}%
}
\end{table}

Image preprocessing procedure is performed to address a few limitations of the dataset. Specifically, the white background occupies much image space in each image, and rice leaf samples are collected with different angles. Therefore, we first applied Otsu's thresholding approach to detect the leaf and cropped out the white background. Subsequently, we adjusted all the samples horizontally aligned. This procedure is described in Figure \ref{Figure 2}. Step 1 includes Otsu's threshold and opening morphology that would detect the shape of the rice leaf. Step 2 refers to removing the unrelated background with the detected object and readjust the image position horizontally. In this way, useless information in the original samples is maximally excluded, which is beneficial for the learning performance. All the rice leaf images are then resized to 299$\times$299 pixels to fix the deep learning model input size. The influence of different input size are also discussed in classification performance section.

\begin{table}[]
\centering
\caption{Classification performance for each class with different models in training procedure}
\label{tab:each_class}
\resizebox{0.65\textwidth}{!}{%
\begin{tabular}{ccccc}
\hline
\multirow{2}{*}{Model} & \multicolumn{4}{c}{Classification category}                                                        \\
                       & Brown Spot              & Healthy                & Rice hispa damage             & Leaf blast             \\ \hline
VGG16                  & 0.712 (0.414)           & 0.629 (0.488)          & 0.600 (0.548)            & 0.563 (0.519)          \\
ResNet50               & 0.764 (0.173)           & 0.632 (0.314)          & 0.533 (0.232)          & 0.359 (0.278)          \\
DenseNet121            & 0.684 (0.282)           & 0.770 (0.275)           & 0.743 (0.167)          & 0.784 (0.190)           \\
MobileNet              & 0.825(0.197)            & 0.937 (0.048)          & \textbf{0.931 (0.062)} & 0.704 (0.243)          \\
InceptionV3            & 0.647 (0.185)           & 0.887 (0.097)          & 0.563 (0.272)          & 0.540 (0.306)           \\
Xception               & 0.856 (0.110)            & 0.954 (0.026)          & 0.824 (0.110)           & 0.702 (0.294)          \\
\textbf{ADSNN-BO}              & \textbf{0.980 (0.029)} & \textbf{0.967 (0.075)} & 0.911 (0.097)          & \textbf{0.937 (0.062)} \\ \hline
\end{tabular}%
}
\end{table}

\begin{figure}[!h]
\captionsetup[subfigure]{justification=centering}
  \centering
  \begin{minipage}[b]{0.47\textwidth}
    \includegraphics[width=\textwidth]{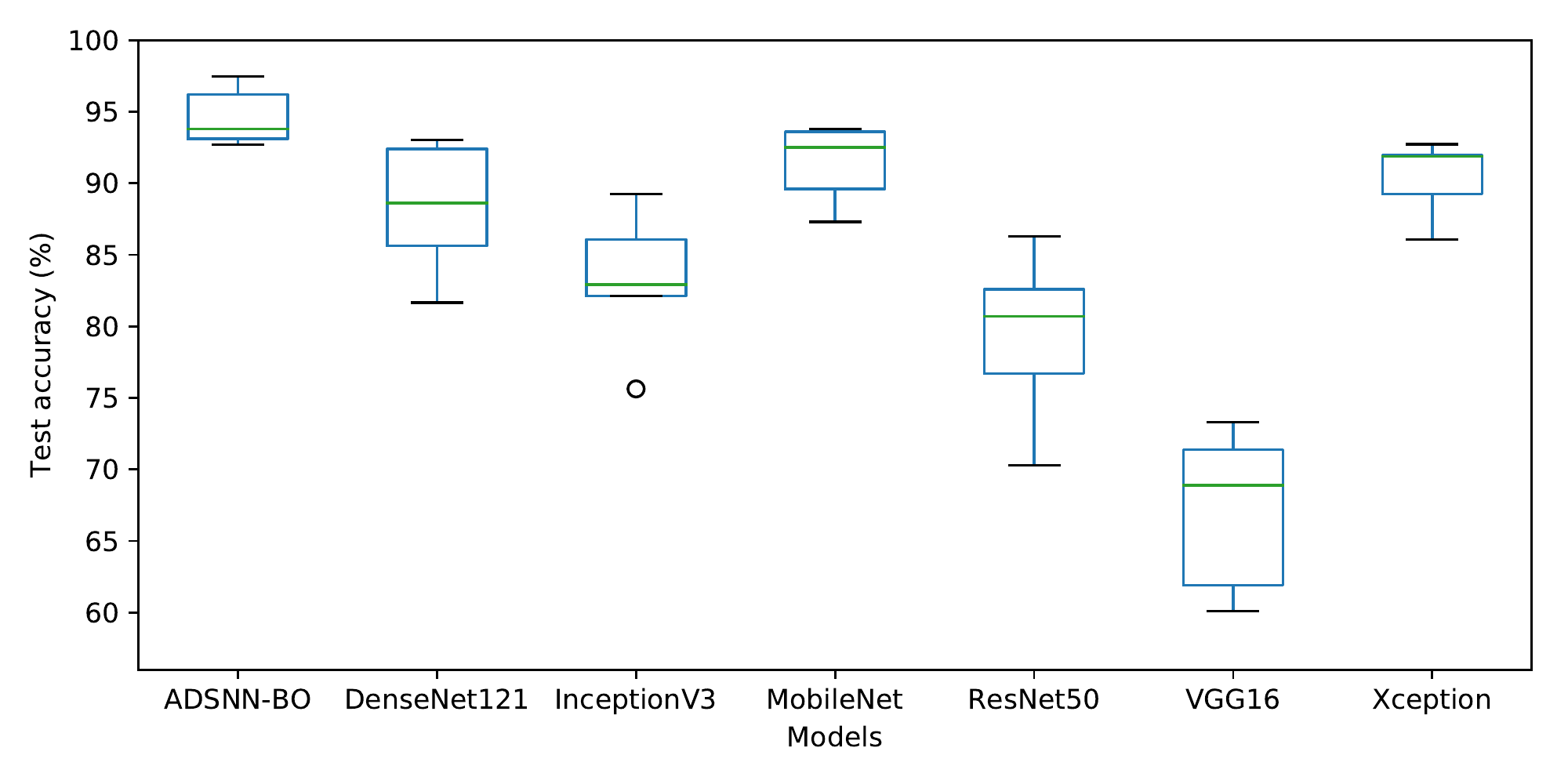}
    \subcaption{Test accuracy}
  \end{minipage}
  \hfill
  \begin{minipage}[b]{0.47\textwidth}
    \includegraphics[width=\textwidth]{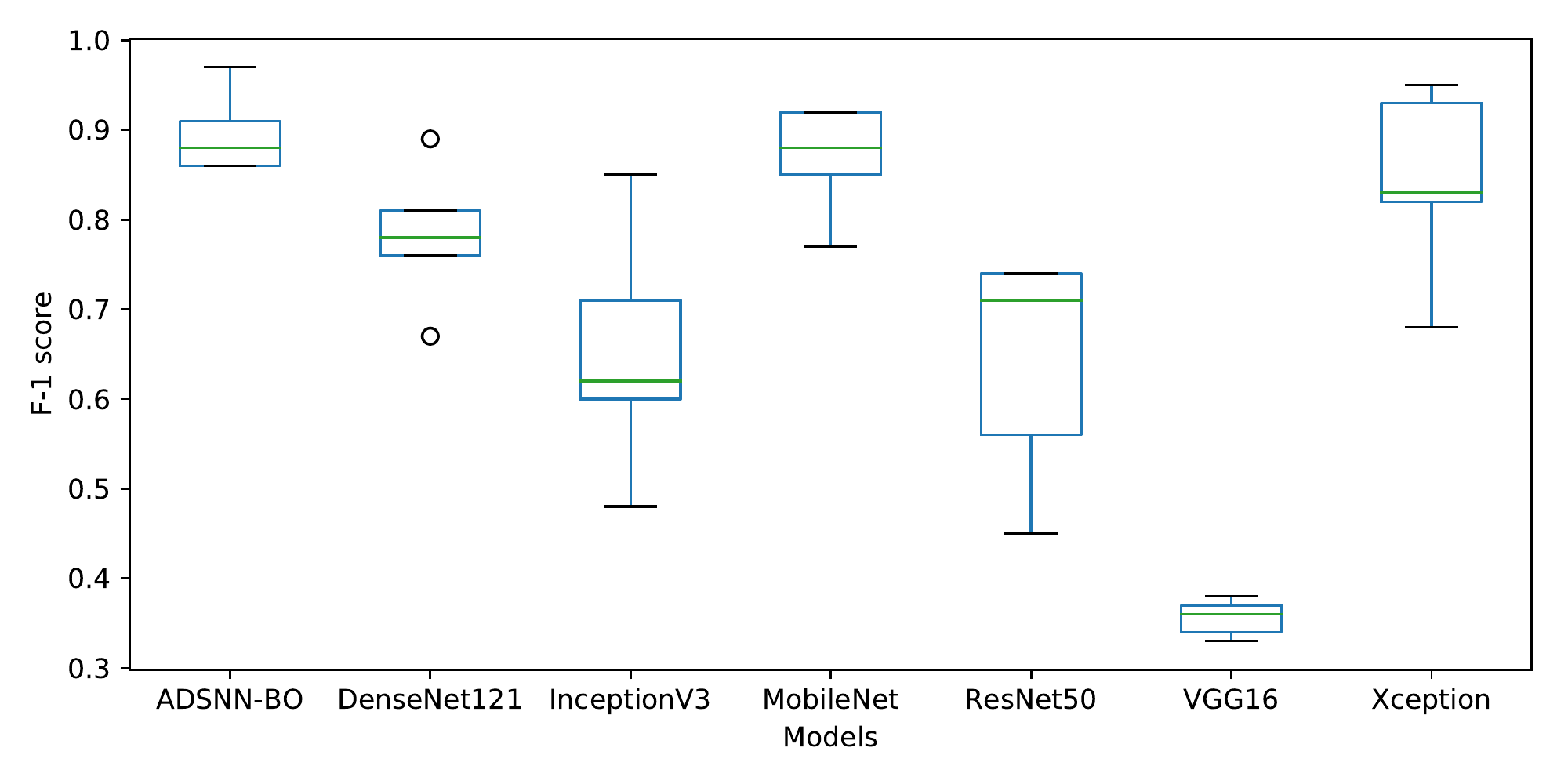}
    \subcaption{F-1 score}
  \end{minipage}
  \begin{minipage}[b]{0.47\textwidth}
    \includegraphics[width=\textwidth]{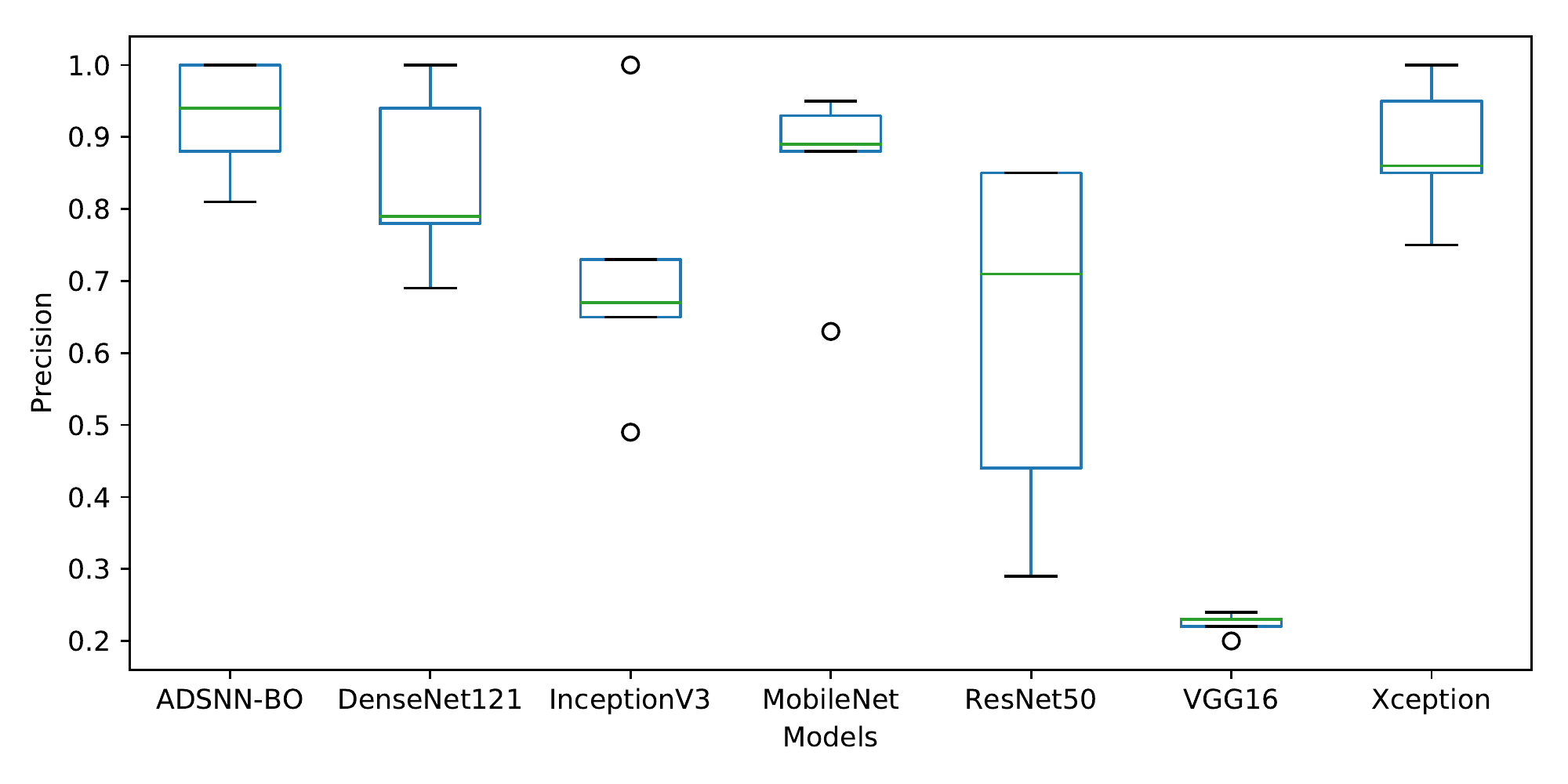}
    \subcaption{Precision}
  \end{minipage}
  \begin{minipage}[b]{0.47\textwidth}
    \includegraphics[width=\textwidth]{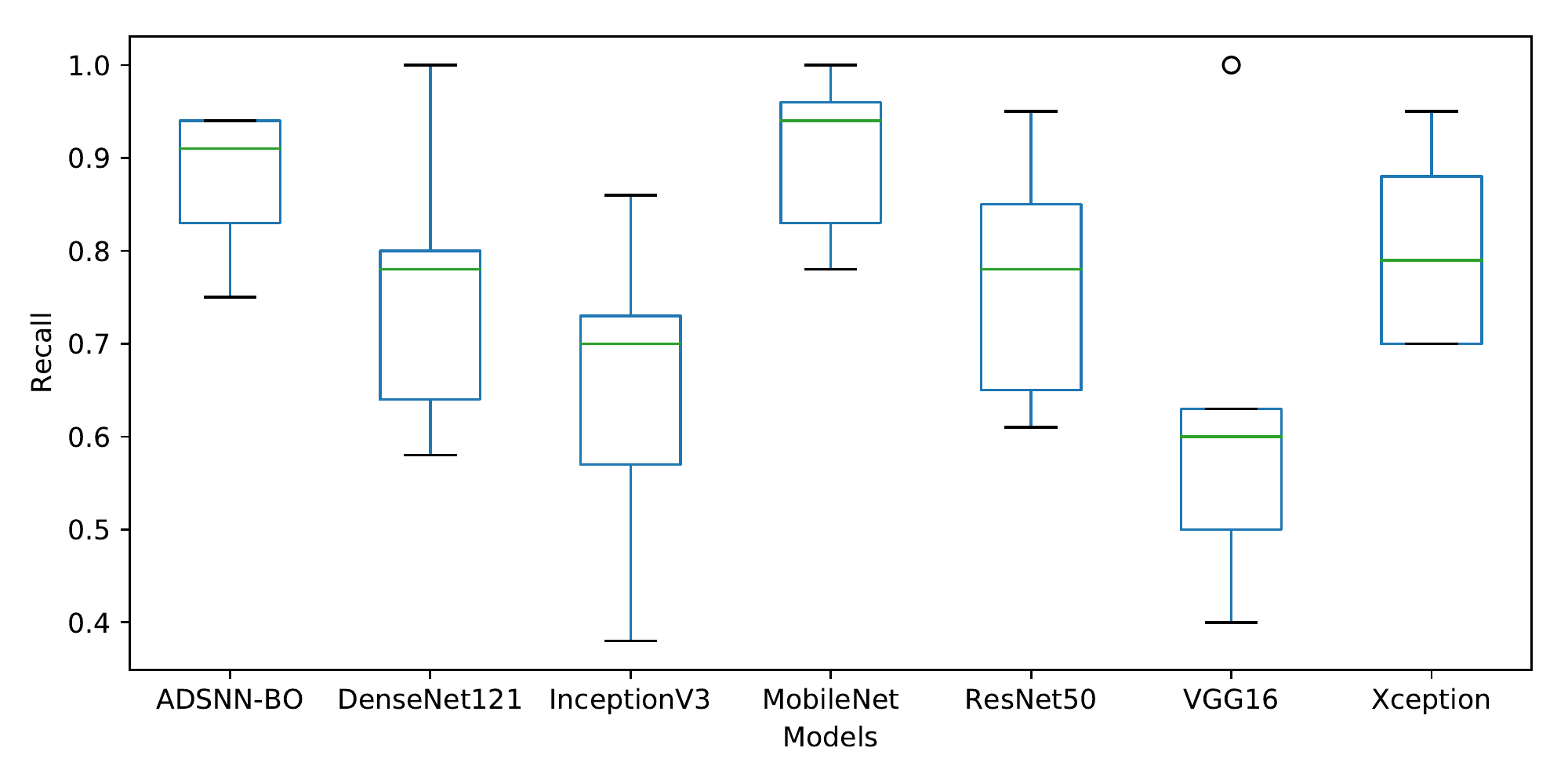}
    \subcaption{Recall}
  \end{minipage}
  \begin{minipage}[b]{0.47\textwidth}
    \includegraphics[width=\textwidth]{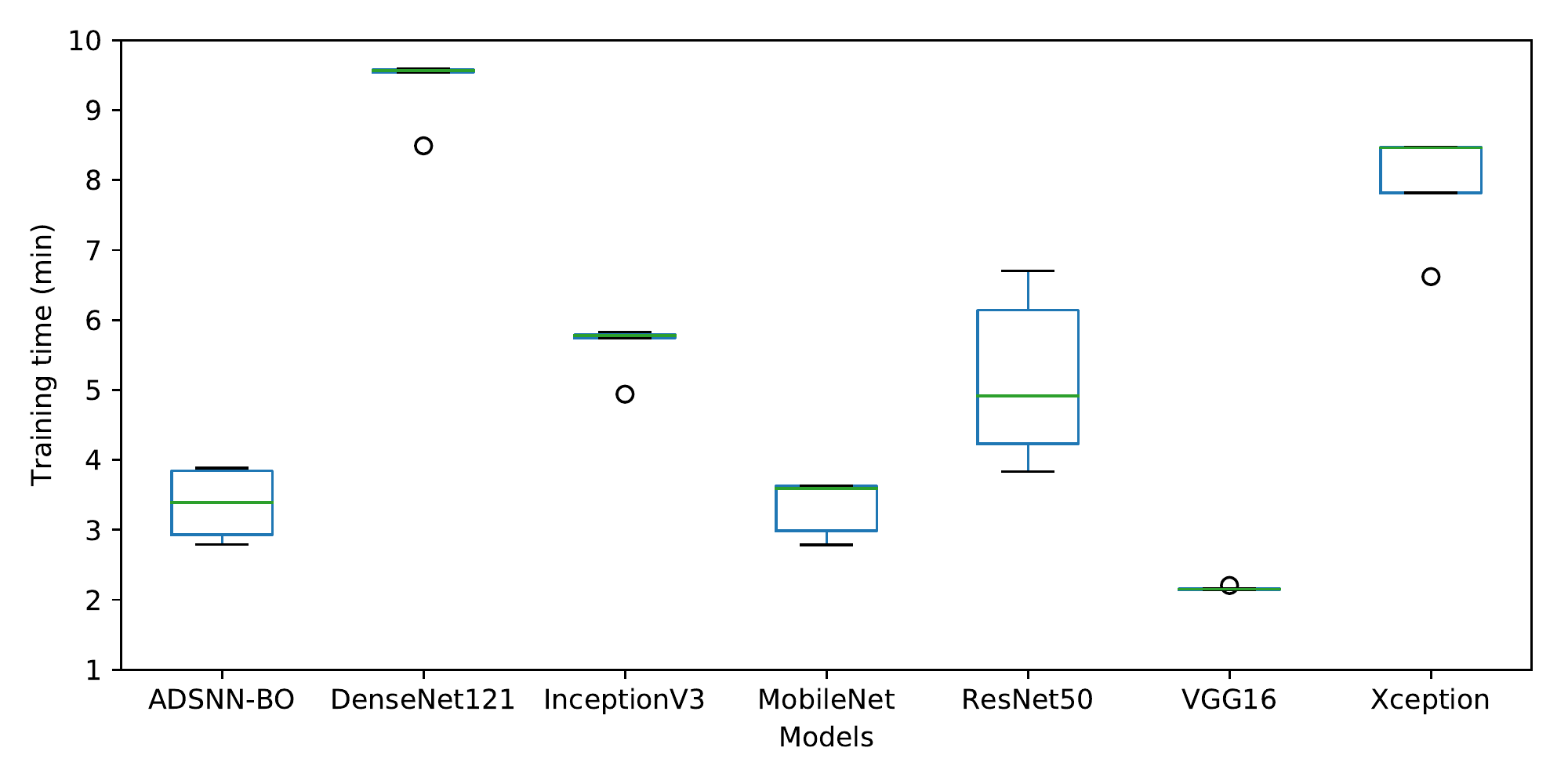}
    \subcaption{Training time}
  \end{minipage}
  \caption{Performance comparison of different deep leaning models}
  \label{Figure 3}
\end{figure}

\begin{figure}[!h]
\captionsetup[subfigure]{justification=centering}
  \centering
  \begin{minipage}[b]{0.47\textwidth}
    \includegraphics[width=\textwidth]{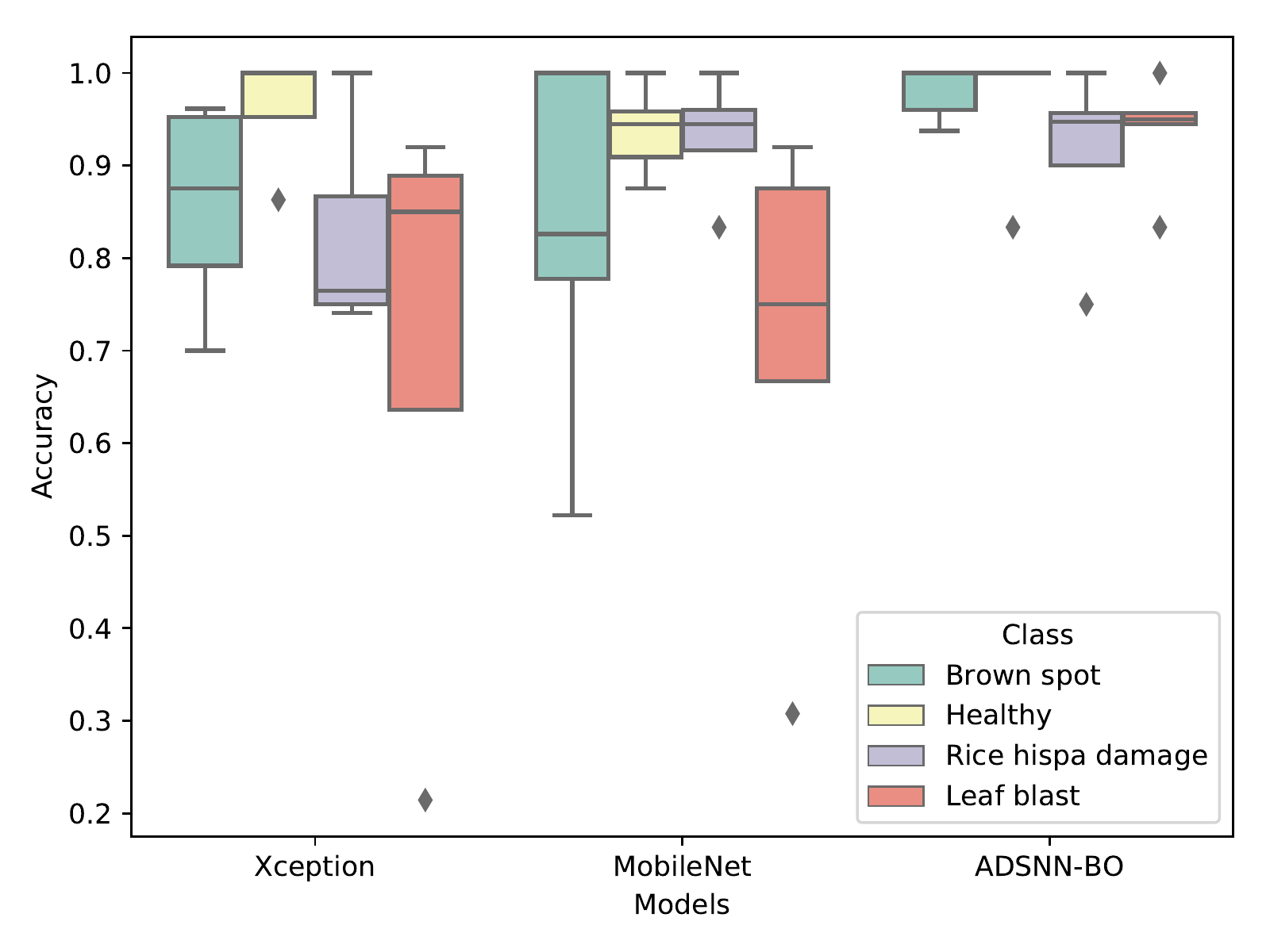}
    \subcaption{Accuracy for different models}
  \end{minipage}
  \begin{minipage}[b]{0.47\textwidth}
    \includegraphics[width=\textwidth]{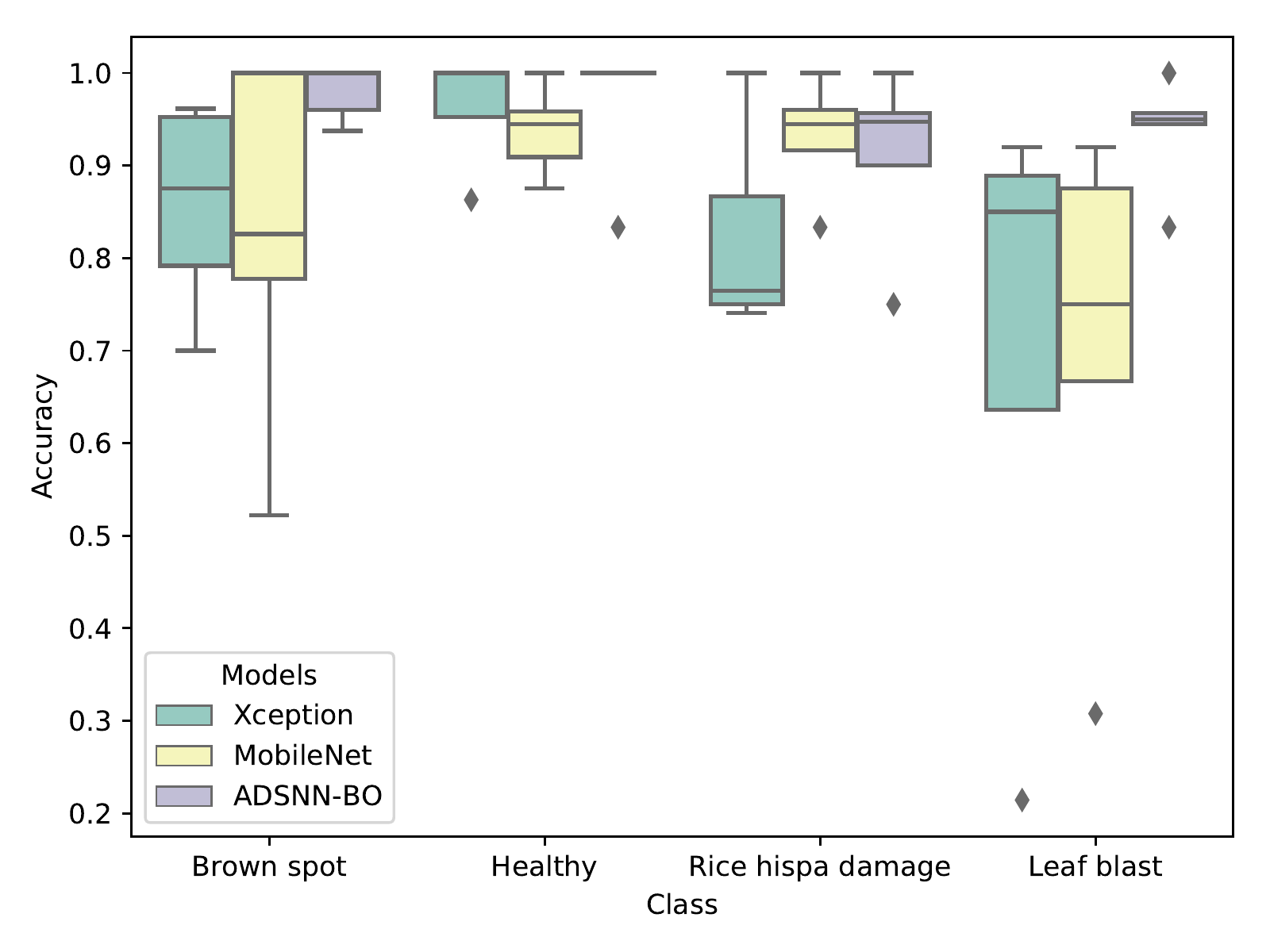}
    \subcaption{Accuracy for different categories}
  \end{minipage}
      \begin{minipage}[b]{0.47\textwidth}
    \includegraphics[width=\textwidth]{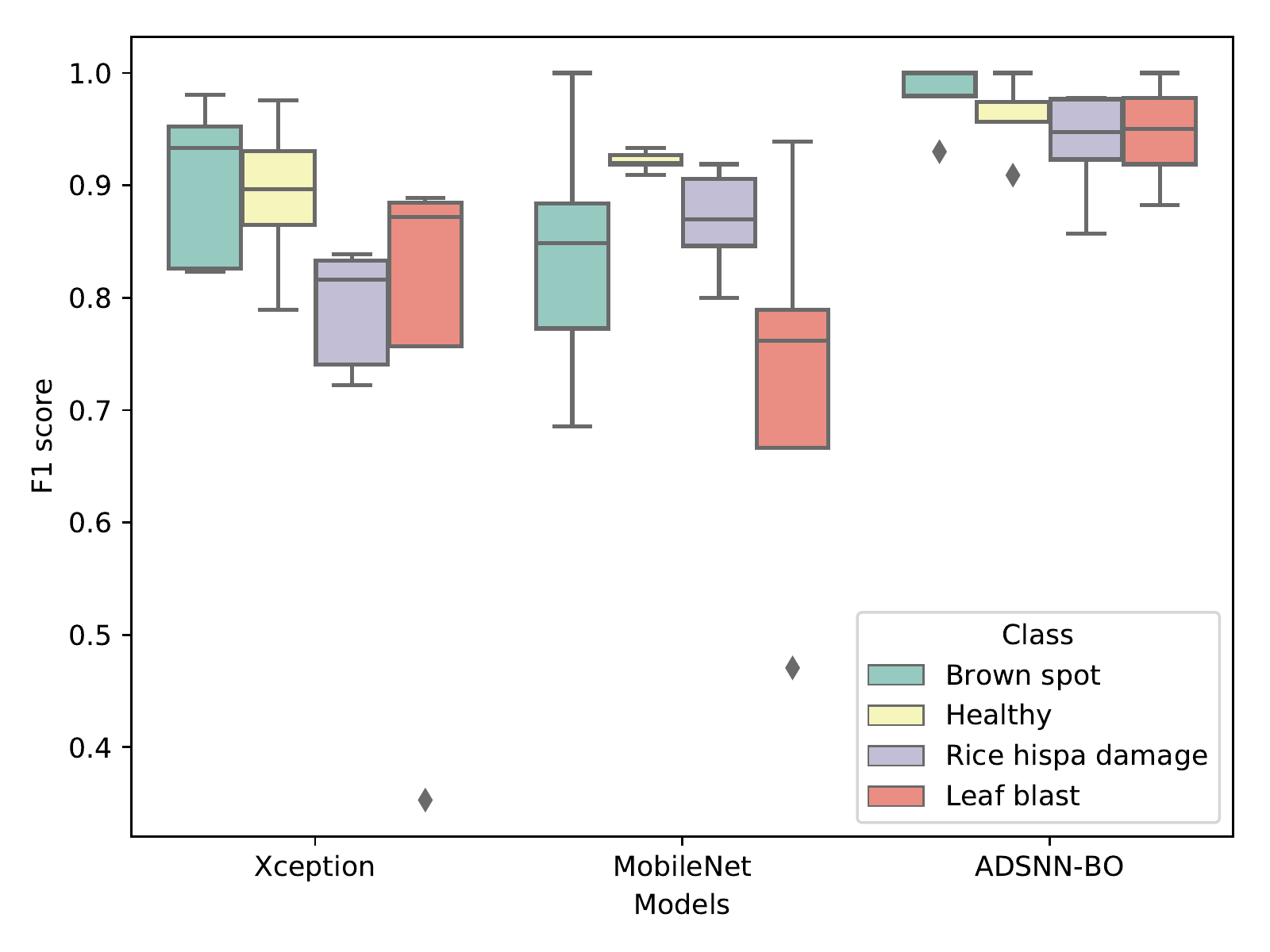}
    \subcaption{F-1 score for different models}
  \end{minipage}
  \begin{minipage}[b]{0.47\textwidth}
    \includegraphics[width=\textwidth]{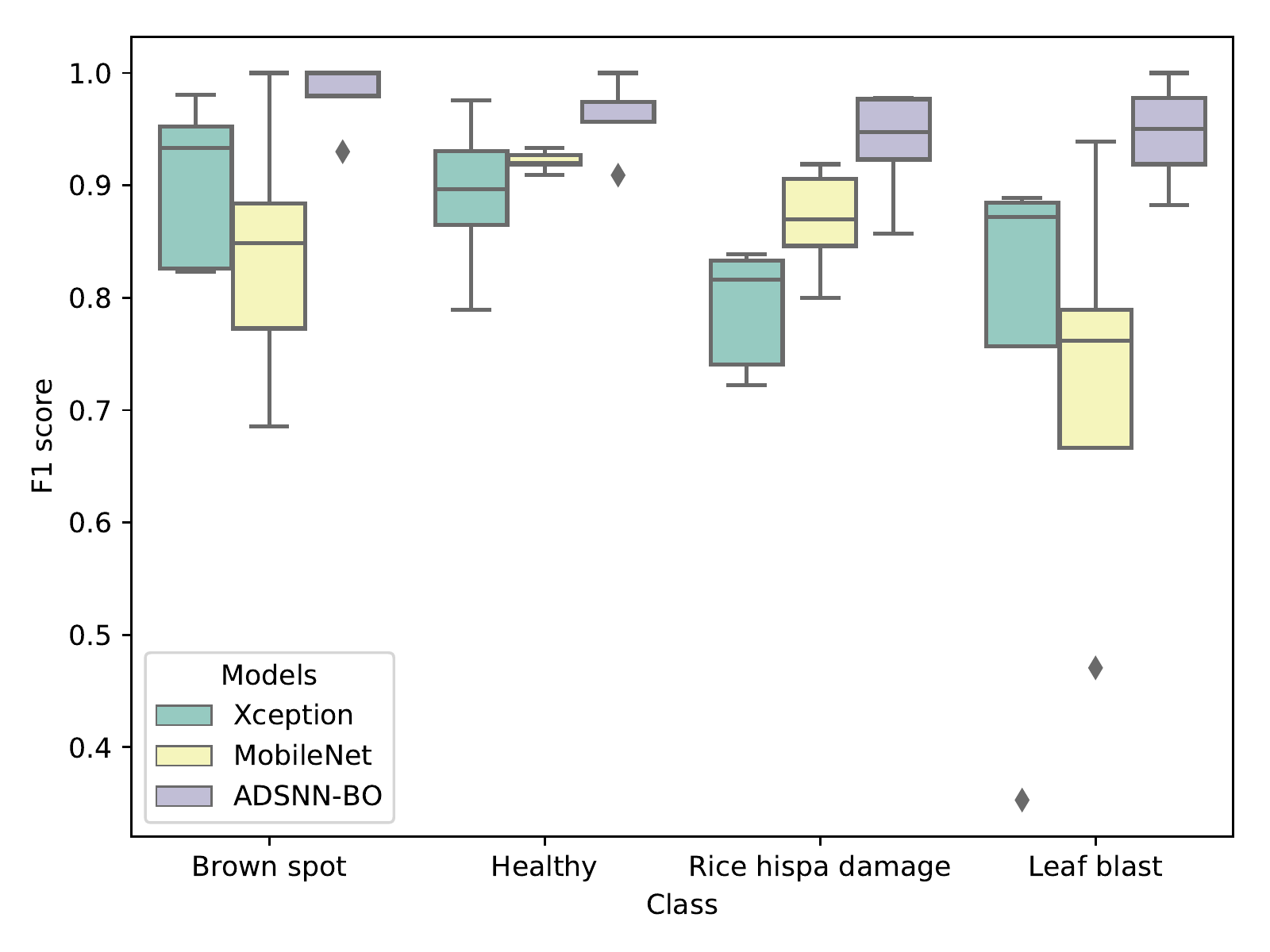}
    \subcaption{F-1 score for different categories}
  \end{minipage}
  \caption{Performance comparison of the top three accuracy models for different classes classification}
  \label{Figure 4}
\end{figure}

\subsection{Performance Measurement}

The proposed ADSNN-BO model is applied to rice disease classification problem. In our experiment, we use dataset consisting of 400 images that are labeled 4 classes of disease categories. The samples are resized to 224 $\times$ 224 as the inputs of the model. Different original deep learning models are tested, including VGG16, ResNet50, DenseNet121, MobileNetV1, Inception V3, and Xception model. We follow the original architecture and test each CNN model with five fold cross-validation. Pre-trained ImageNet weights are utilized to achieve better performance. For each training set, we further split it into training and validation set with the ratio of 0.7 and 0.3. The number of running epoch is set to be 100. All experiments are performed using TensorFlow under the computational specification of 64-bit Windows 10, with Intel i9-9900
processor (3.10GHz), 32GB random-access memory (RAM), and NVIDIA GeForce RTX 2080.

\begin{equation} \label{eq8}
\begin{aligned}
p_{i} =\frac{n_{ii}}{\sum _{j=1}^{M}n_{ji}}\end{aligned}
\end{equation}

\begin{equation} \label{eq9}
\begin{aligned}
r_{i} =\frac{n_{ii}}{\sum _{j=1}^{M}n_{ij}}\end{aligned}
\end{equation}

\begin{equation} \label{eq10}
\begin{aligned}
f_{i} =\frac{2r_{i}p_{i}}{r_{i}+p_{i}}\end{aligned}
\end{equation}

To measure the model classification performance, a multi-class confusion matrix is introduced \citep{wang2019deep}. Given a problem with M classes, a $M \times M$ matrix is constructed, where $n_{ij}$ denotes the number of samples with the actual label $i$ that are classified as the predicted label $j$. Precision, recall, and F1-score of each class can be calculated based on the returned confusion matrix. Those measurement equations are indicated in Eq.[\ref{eq8}]-[\ref{eq10}]. 
Although Matthews correlation coefficient is identified to perform better for binary classification, it is not suitable for our multi-class classification tasks\citep{chicco2020advantages}. The experimental results evaluated based on the performance measurement are shown in Table \ref{tab:performance}. The table reveals the mean value and standard deviation of each measurement based on five fold cross-validation results. The bold characters indicate the best performance among the tested deep learning models.

According to the comparisons of the model performance, our proposed ADSNN-BO model achieves the highest test accuracy along with the highest precision and F-1 score. The original MobileNet model attains the highest recall value. The training time of the proposed model also prevails most of the models, which is close to the original MobileNet. ADSNN-BO increased the precision of 8.18\%, F-1 score of 3.23\%, and the accuracy of 3.6\% compared with the original MobileNet model. Furthermore, we conduct performance comparisons of the top three accuracy models, which are proposed ADSNN-BO, MobileNet, and Xception. Boxplots of accuracy and F-1 score are shown in Figure \ref{Figure 4}. In general, for each type of rice disease, ADSNN-BO outperforms other deep learning models. According to the comparisons of different category performance within one model, it can be considered that leaf blast is the most difficult disease to detect and classify. Further explanation will be discussed through filter visualization.

\begin{figure}[!h]
\captionsetup[subfigure]{justification=centering}
  \centering
  \includegraphics[width=0.8\textwidth]{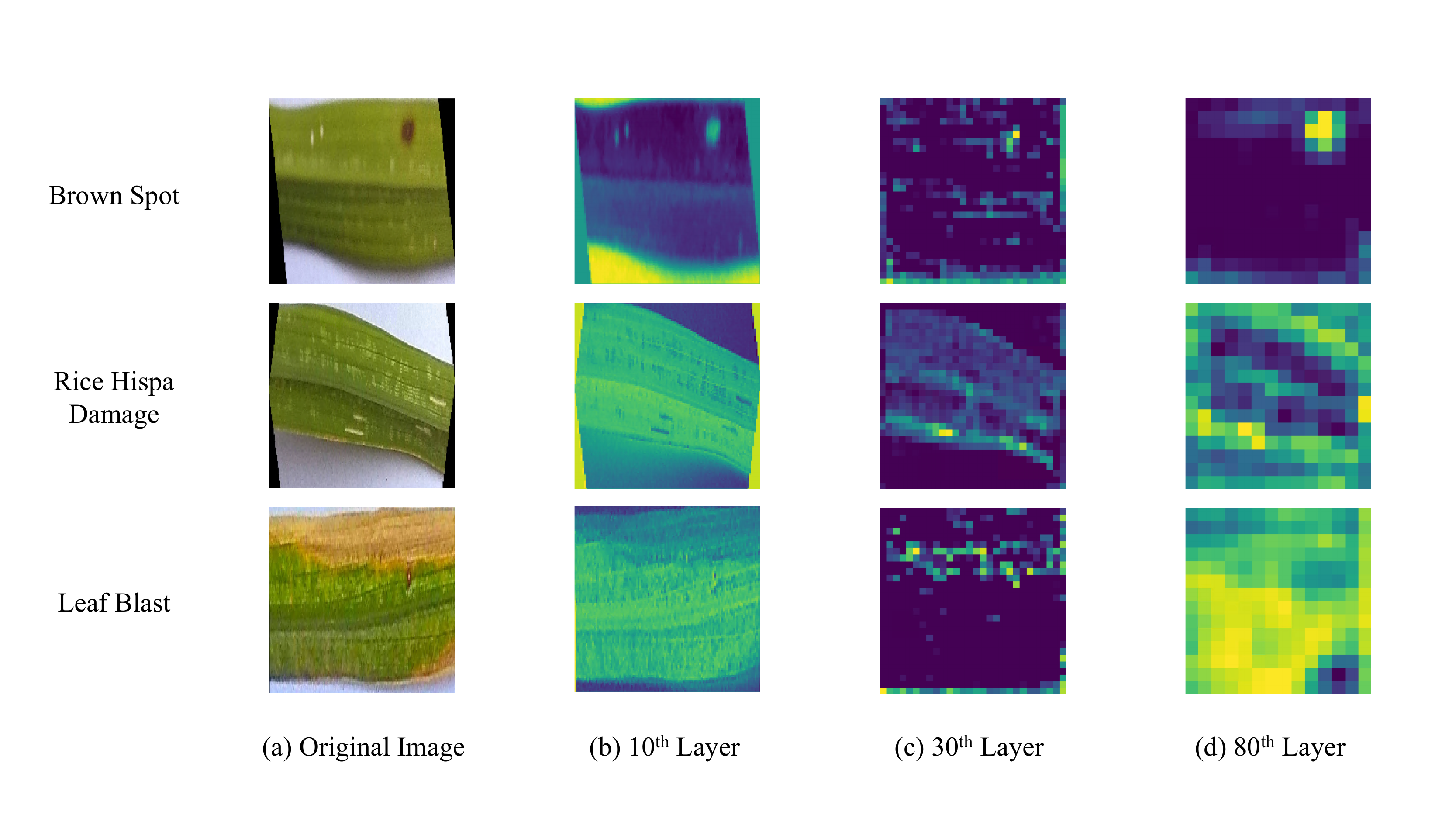}
  \caption{Activation maps of different layers for each type of disease}
  \label{Figure 5}
\end{figure}

\begin{algorithm}[!h]
\SetAlgoLined
 Start from a gray image with some noise.\\
 Build a loss function that maximizes the activation of the $n$th filter of the layer under consideration.\\
 Compute the gradient of the input picture with regard to this loss.\\
 Normalize the gradient.\\
 Return the loss and gradients given the input picture.\\
  Run gradient ascent for $i$ steps.\\
  Utility function to convert a tensor into a valid image to get the output.\\
 \caption{Convnet Filter Visualization Algorithm}
 \label{tab:algorithm2}
\end{algorithm}

\begin{figure}[!h]
\captionsetup[subfigure]{justification=centering}
  \centering
  \begin{minipage}[b]{\textwidth}
    \includegraphics[width=\textwidth]{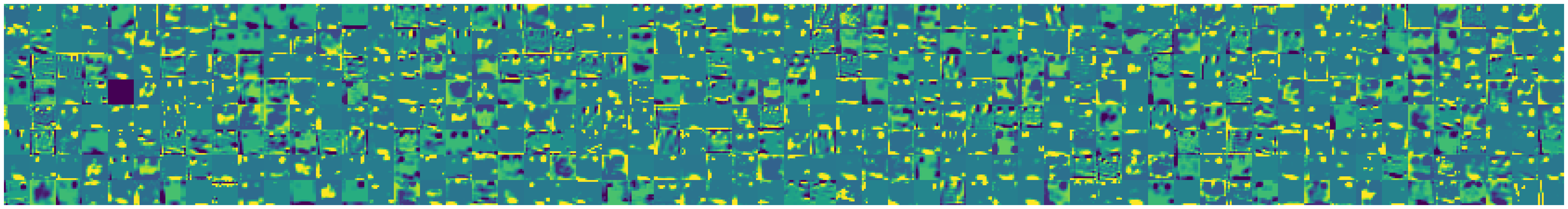}
    \subcaption{Brown spot}
  \end{minipage}
  \begin{minipage}[b]{\textwidth}
    \includegraphics[width=\textwidth]{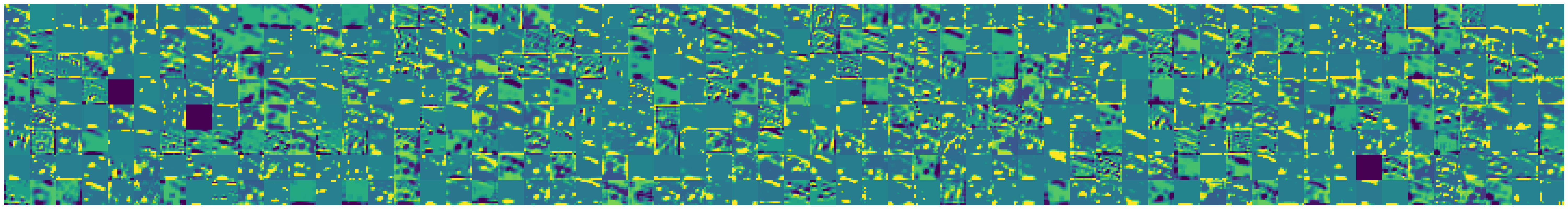}
    \subcaption{Rice hispa damage}
  \end{minipage}
  \begin{minipage}[b]{\textwidth}
    \includegraphics[width=\textwidth]{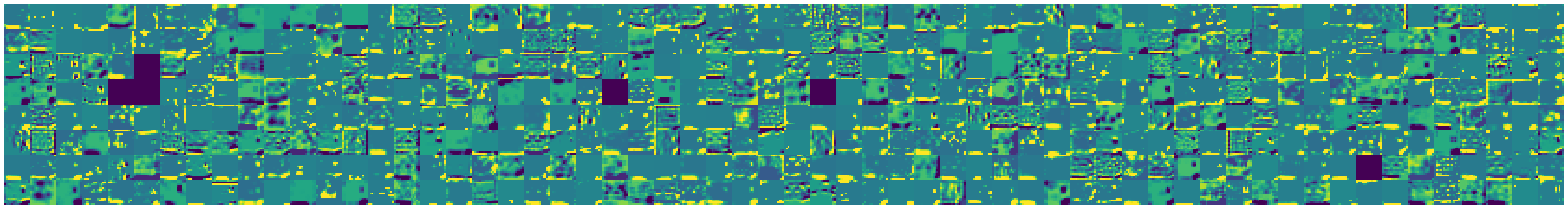}
    \subcaption{Leaf blast}
  \end{minipage}
  \caption{Activation maps of one single layer (70th) in the model for each type of disease}
  \label{Figure 6}
\end{figure}

\begin{figure}[!h]
\captionsetup[subfigure]{justification=centering}
  \begin{minipage}[b]{\textwidth}
  \centering
    \includegraphics[width=0.85\textwidth]{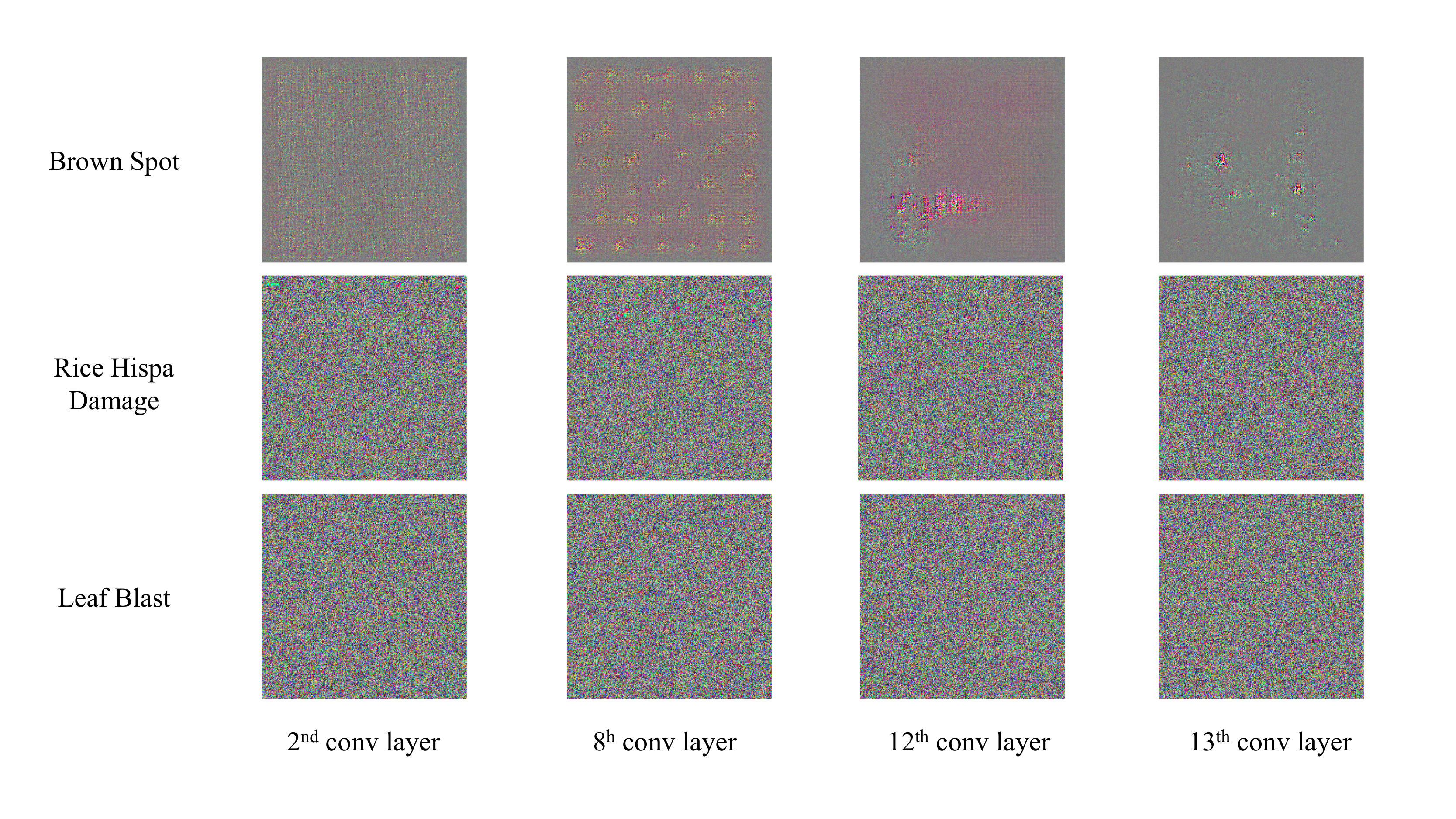}
    \subcaption{Original MobileNet model}
  \end{minipage}
  \\
  \begin{minipage}[b]{\textwidth}
    \centering
    \includegraphics[width=0.85\textwidth]{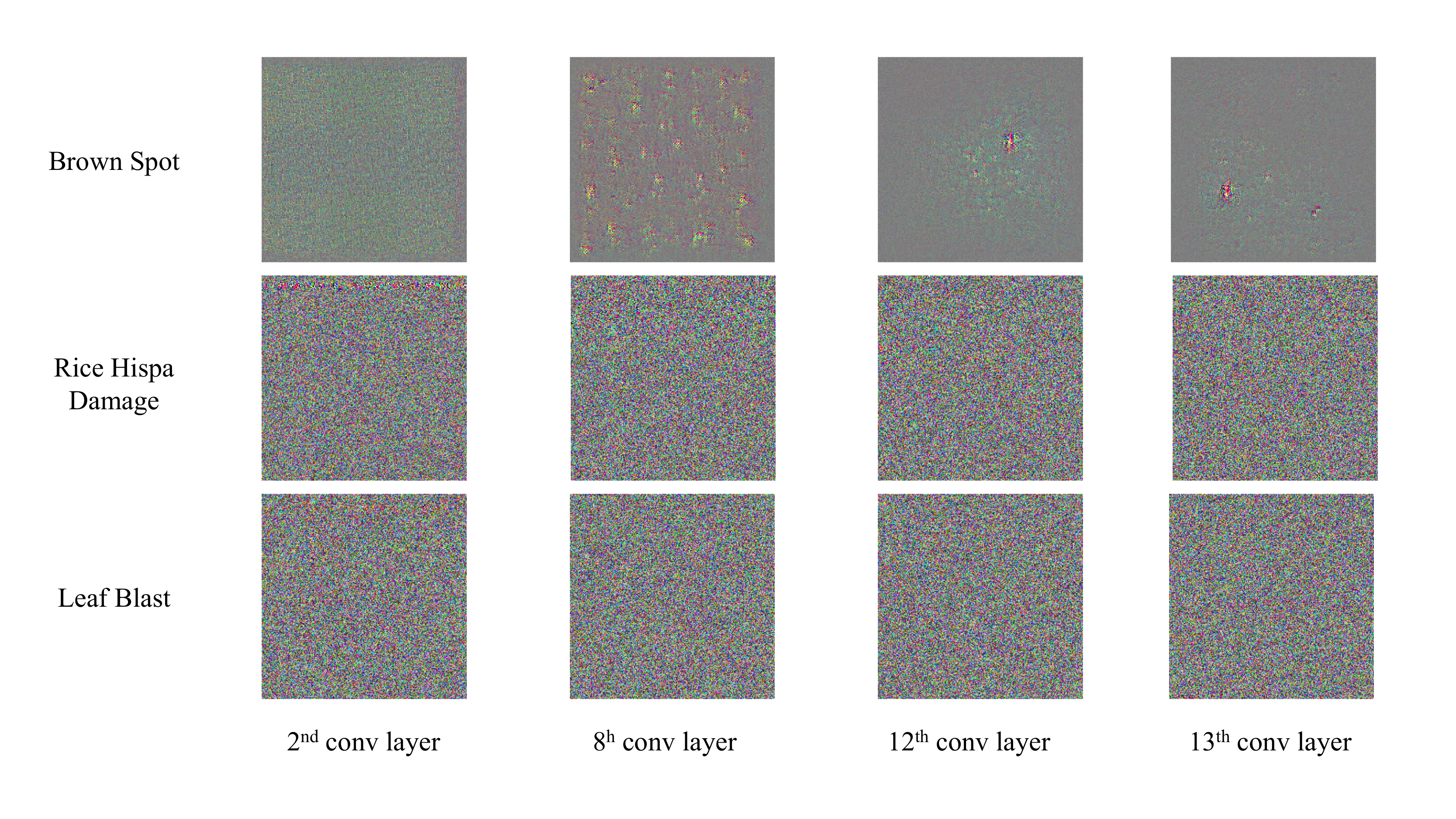}
    \subcaption{ADSNN-BO model}  
  \end{minipage}
  \caption{Filters visualization comparison of different convolutional layer (2nd, 8th, 12th, 13th depthwise layer) for the three rice diseases}
  \label{Figure 7}
\end{figure}

\subsection{Feature Evaluation}

Features that deep learning models extract are evaluated based on two major approaches: (1) activation map, and (2) filters visualization. Activation map, considered as the most straight-forward visualization technique, will provide a visual representation of the activation numbers within different layers of the neural network. In order to generate a series of activation maps throughout running the model, we first load and fix the weights of entire network with the best performed model. Then an original rice image sample is input into the model for testing. After each layer's operation, the output will tell what type of input maximally activates the layer. A series of selected activation maps are generated for our proposed ADSNN-BO model in Figure \ref{Figure 5}. The figures reveal how the model deals with different rice diseases. For example, it is shown that the network captures and extracts the spot pattern successfully when detecting brown spot disease. In addition, activation maps of all the filters within one layer are presented in Figure \ref{Figure 6}. According to the results, it is apparent that most of the filters are getting activated with regard to the pattern for each disease, and various types of patterns are activated for different disease categories. The blank convolution outputs mean that the pattern encoded by the filters were not found in the input image. These patterns must be complex shapes most likely that are not present in the input image.


\begin{figure}[!h]
\captionsetup[subfigure]{justification=centering}
  \begin{minipage}[b]{\textwidth}
  \centering
    \includegraphics[width=0.2\textwidth]{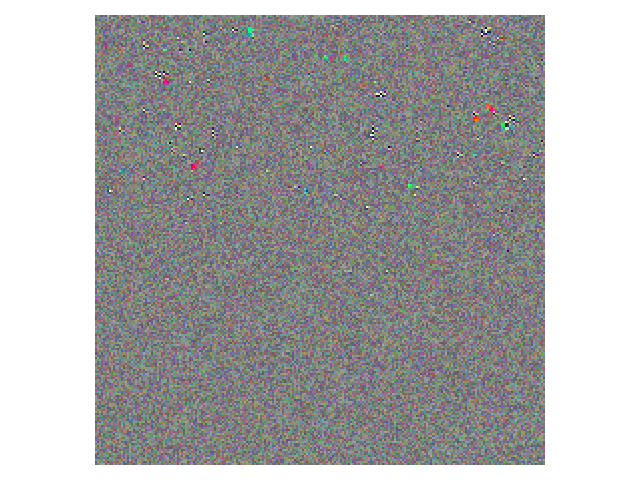}
    \includegraphics[width=0.2\textwidth]{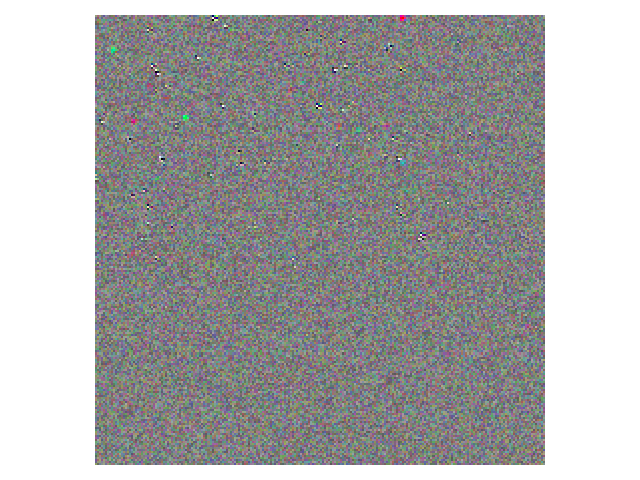}
    \includegraphics[width=0.2\textwidth]{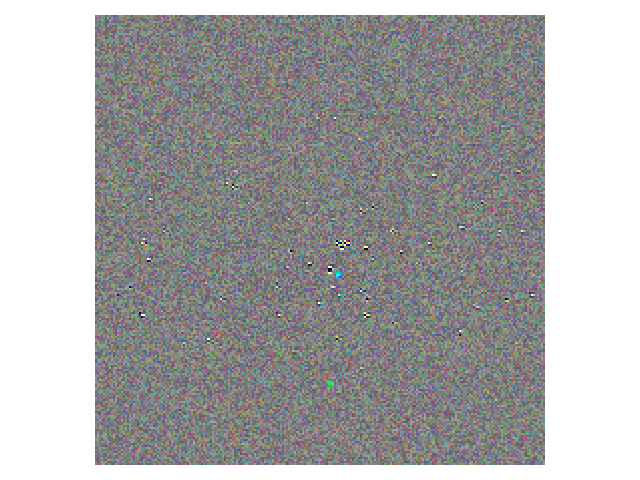}
    \subcaption{Original MobileNet model}
  \end{minipage}
  \\
  \begin{minipage}[b]{\textwidth}
    \centering
    \includegraphics[width=0.2\textwidth]{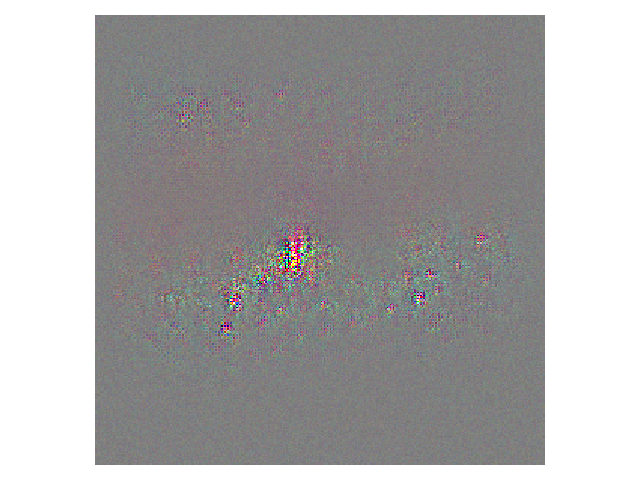}
    \includegraphics[width=0.2\textwidth]{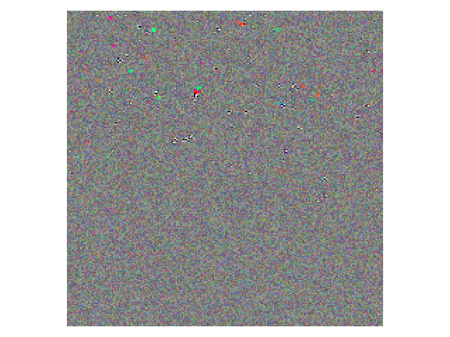}
    \includegraphics[width=0.2\textwidth]{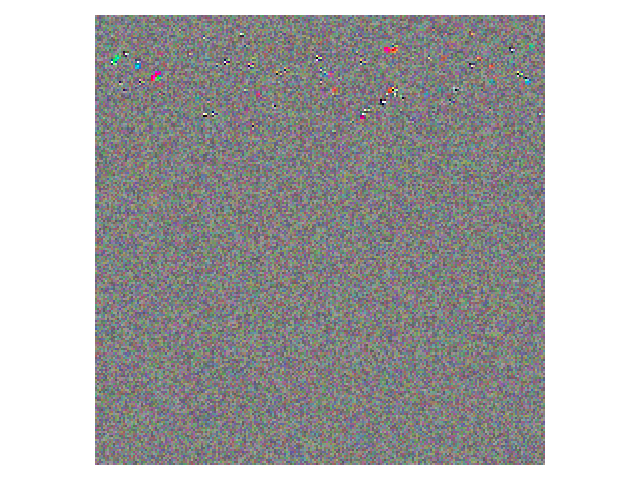}
    \subcaption{ADSNN-BO model}  
  \end{minipage}
  \caption{Visualizations of the last convolutional layer for each type of disease, brown spot, rice hispa damage, leaf blast, for original MobileNet and ADSNN-BO model, respectively.}
  \label{Figure 8}
\end{figure}

Different from activation maps, filter visualization of the model is able to show the filters which are the weights that a model learns. Here, we compare the original MobilNet with our proposed ADSNN-BO model using filters visualization. This visualization procedure starts from a random image, repeatedly apply the gradient ascent, and convert the resulting input image back to a displayable form. The detailed algorithm is shown in Algorithm \ref{tab:algorithm2}. The comparison result is shown in Figure \ref{Figure 7} - \ref{Figure 8}. Figure \ref{Figure 7} indicates the filters of different convolutional layers addressing three rice diseases in two models. In this figure, we can conclude that top layers are usually informative to show the learning ability of a model. Because the lower portion of the model constructure often captures general pattern and information of an image which is meaningless. The filters belonged to top layers are commonly comprehensible and able to evaluate a model's performance. Additionally, we can state that for one typical disease such as brown spot, both models are responsive to the spot pattern. While the filters in ADSNN-BO model are more recognizable and possess less noise compared with the original MobileNet. Visualizations of the last layer for different disease are shown in Figure \ref{Figure 8}. ADSNN-BO model outperforms original MobileNet when detecting brown spot pattern as the classification accuracy results indicate in Table \ref{tab:algorithm}. The results of other disease categories are similar according to the visualization.

\section{Conclusions}

In this paper, an overview of deep learning models and implementations in rice diseases is provided. Six existing deep learning models have been implemented and compared with each other for rice disease classification problem. Pre-trained ImageNet weights are used to provide better results for all CNN architectures. A new model named ADSNN-OB has been proposed,tested, and discussed. The details of data preparation steps are introduced. Various experiments are conducted which show that our proposed ADSNN-OB model outperforms other deep learning models in various scenarios. Feature evaluation including activation map and filters visualization is also implemented to illustrate the performance of the model. The results show that a new model which outperforms all tested state-of-the-art models and fits well with the mobile device environment is created.

In the next step, we plan to explore more on our proposed ADSNN-OB model with different optimization methods as well as hyperparameter tuning to improve the performance and make the procedure more effectively. Also, since the dataset we used is different from those in the earlier literature, we tend to apply the model on different public datasets. Integrating location, weather and soil information along with the given rice disease image can be interesting to investigate for a more comprehensive and automated rice disease detection and classification mechanism.


\bibliographystyle{apalike}\biboptions{authoryear}
\bibliography{mybibfile.bib}

\begin{thebibliography}{}

\bibitem[Tom, 2020]{Tomato}
 (2020).
\newblock Attention embedded residual cnn for disease detection in tomato
  leaves.
\newblock {\em Applied Soft Computing}, 86:105933.

\bibitem[Agrios, 2005]{agrios2005plant}
Agrios, G. (2005).
\newblock Plant pathology 5th edition: Elsevier academic press.
\newblock {\em Burlington, Ma. USA}, pages 79--103.

\bibitem[Bello et~al., 2019]{bello2019attention}
Bello, I., Zoph, B., Vaswani, A., Shlens, J., and Le, Q.~V. (2019).
\newblock Attention augmented convolutional networks.
\newblock In {\em Proceedings of the IEEE International Conference on Computer
  Vision}, pages 3286--3295.

\bibitem[Chen et~al., 2019]{chen2019ricetalk}
Chen, W.-L., Lin, Y.-B., Ng, F.-L., Liu, C.-Y., and Lin, Y.-W. (2019).
\newblock Ricetalk: Rice blast detection using internet of things and
  artificial intelligence technologies.
\newblock {\em IEEE Internet of Things Journal}, 7(2):1001--1010.

\bibitem[Chicco and Jurman, 2020]{chicco2020advantages}
Chicco, D. and Jurman, G. (2020).
\newblock The advantages of the matthews correlation coefficient (mcc) over f1
  score and accuracy in binary classification evaluation.
\newblock {\em BMC genomics}, 21(1):6.

\bibitem[Chollet, 2017]{chollet2017xception}
Chollet, F. (2017).
\newblock Xception: Deep learning with depthwise separable convolutions.
\newblock In {\em Proceedings of the IEEE conference on computer vision and
  pattern recognition}, pages 1251--1258.

\bibitem[Chung et~al., 2016]{chung2016detecting}
Chung, C.-L., Huang, K.-J., Chen, S.-Y., Lai, M.-H., Chen, Y.-C., and Kuo,
  Y.-F. (2016).
\newblock Detecting bakanae disease in rice seedlings by machine vision.
\newblock {\em Computers and electronics in agriculture}, 121:404--411.

\bibitem[Duong-Trung et~al., 2019]{duong2019classification}
Duong-Trung, N., Quach, L.-D., Nguyen, M.-H., and Nguyen, C.-N. (2019).
\newblock Classification of grain discoloration via transfer learning and
  convolutional neural networks.
\newblock In {\em Proceedings of the 3rd International Conference on Machine
  Learning and Soft Computing}, pages 27--32.

\bibitem[E and A., 2018]{Goceri}
E, G. and A., G. (2018).
\newblock On the importance of batch size for deep learning.
\newblock {\em Int Conf on Mathematics (ICOMATH2018), An Istanbul Meeting for
  World Mathematicians, Istanbul, Turkey}.

\bibitem[Esen et~al., 2017]{esen2017modelling}
Esen, H., Esen, M., and Ozsolak, O. (2017).
\newblock Modelling and experimental performance analysis of solar-assisted
  ground source heat pump system.
\newblock {\em Journal of Experimental \& Theoretical Artificial Intelligence},
  29(1):1--17.

\bibitem[Esen et~al., 2008]{esen2008performance}
Esen, H., Inalli, M., Sengur, A., and Esen, M. (2008).
\newblock Performance prediction of a ground-coupled heat pump system using
  artificial neural networks.
\newblock {\em Expert Systems with Applications}, 35(4):1940--1948.

\bibitem[Esen et~al., 2009]{esen2009artificial}
Esen, H., Ozgen, F., Esen, M., and Sengur, A. (2009).
\newblock Artificial neural network and wavelet neural network approaches for
  modelling of a solar air heater.
\newblock {\em Expert systems with applications}, 36(8):11240--11248.

\bibitem[Francois, 2017]{francois2017deep}
Francois, C. (2017).
\newblock Deep learning with python.

\bibitem[Frazier, 2018]{frazier2018tutorial}
Frazier, P.~I. (2018).
\newblock A tutorial on bayesian optimization.
\newblock {\em arXiv preprint arXiv:1807.02811}.

\bibitem[Goceri, 2019a]{Goceri2}
Goceri (2019a).
\newblock Analysis of deep networks with residual blocks and different
  activation functions: Classification of skin diseases.
\newblock In {\em 2019 Ninth International Conference on Image Processing
  Theory, Tools and Applications (IPTA)}, pages 1--6.

\bibitem[Goceri, 2019b]{Goceri3}
Goceri (2019b).
\newblock Challenges and recent solutions for image segmentation in the era of
  deep learning.
\newblock In {\em 2019 Ninth International Conference on Image Processing
  Theory, Tools and Applications (IPTA)}, pages 1--6.

\bibitem[Goceri, 2019c]{goceri2019diagnosis}
Goceri, E. (2019c).
\newblock Diagnosis of alzheimer's disease with sobolev gradient-based
  optimization and 3d convolutional neural network.
\newblock {\em International journal for numerical methods in biomedical
  engineering}, 35(7):e3225.

\bibitem[Goceri, 2019d]{goceri2019skin}
Goceri, E. (2019d).
\newblock Skin disease diagnosis from photographs using deep learning.
\newblock In {\em ECCOMAS Thematic Conference on Computational Vision and
  Medical Image Processing}, pages 239--246. Springer.

\bibitem[Goceri, 2020]{goceri2020deep}
Goceri, E. (2020).
\newblock Deep learning based classification of facial dermatological
  disorders.
\newblock {\em Computers in Biology and Medicine}, 128:104118.

\bibitem[G{\"o}{\c{c}}eri, 2020]{gocceri2020impact}
G{\"o}{\c{c}}eri, E. (2020).
\newblock Impact of deep learning and smartphone technologies in dermatology:
  Automated diagnosis.
\newblock In {\em 2020 Tenth International Conference on Image Processing
  Theory, Tools and Applications (IPTA)}, pages 1--6. IEEE.

\bibitem[Goceri and Karakas, 2020]{goceri4}
Goceri, E. and Karakas, A.~A. (2020).
\newblock Comparative evaluations of cnn based networks for skin lesion
  classification.
\newblock In {\em 14th International Conference on Computer Graphics,
  Visualization, Computer Vision and Image Processing (CGVCVIP), Zagreb,
  Croatia}, pages 1--6.

\bibitem[Howard et~al., 2017]{howard2017mobilenets}
Howard, A.~G., Zhu, M., Chen, B., Kalenichenko, D., Wang, W., Weyand, T.,
  Andreetto, M., and Adam, H. (2017).
\newblock Mobilenets: Efficient convolutional neural networks for mobile vision
  applications.
\newblock {\em arXiv preprint arXiv:1704.04861}.

\bibitem[HuyDo, 2019]{HuyDo2019}
HuyDo (2019).
\newblock Rice diseases image dataset: An image dataset for rice and its
  diseases.

\bibitem[Liang et~al., 2019]{liang2019rice}
Liang, W.-j., Zhang, H., Zhang, G.-f., and Cao, H.-x. (2019).
\newblock Rice blast disease recognition using a deep convolutional neural
  network.
\newblock {\em Scientific reports}, 9(1):1--10.

\bibitem[Lu et~al., 2018]{lu2018dual}
Lu, H., Wang, H., Zhang, Q., Won, D., and Yoon, S.~W. (2018).
\newblock A dual-tree complex wavelet transform based convolutional neural
  network for human thyroid medical image segmentation.
\newblock In {\em 2018 IEEE International Conference on Healthcare Informatics
  (ICHI)}, pages 191--198. IEEE.

\bibitem[Lu et~al., 2017]{lu2017identification}
Lu, Y., Yi, S., Zeng, N., Liu, Y., and Zhang, Y. (2017).
\newblock Identification of rice diseases using deep convolutional neural
  networks.
\newblock {\em Neurocomputing}, 267:378--384.

\bibitem[Rafeed~Rahman et~al., 2018]{rafeed2018identification}
Rafeed~Rahman, C., Saha~Arko, P., Eunus~Ali, M., Khan, M. A.~I., Hasan~Apon,
  S., Nowrin, F., and Wasif, A. (2018).
\newblock Identification and recognition of rice diseases and pests using
  convolutional neural networks.
\newblock {\em arXiv}, pages arXiv--1812.

\bibitem[Ramesh and Vydeki, 2020]{ramesh2020recognition}
Ramesh, S. and Vydeki, D. (2020).
\newblock Recognition and classification of paddy leaf diseases using optimized
  deep neural network with jaya algorithm.
\newblock {\em Information processing in agriculture}, 7(2):249--260.

\bibitem[Rehman et~al., 2019]{rehman2019current}
Rehman, T.~U., Mahmud, M.~S., Chang, Y.~K., Jin, J., and Shin, J. (2019).
\newblock Current and future applications of statistical machine learning
  algorithms for agricultural machine vision systems.
\newblock {\em Computers and electronics in agriculture}, 156:585--605.

\bibitem[Shrivastava et~al., 2019]{shrivastava2019rice}
Shrivastava, V.~K., Pradhan, M.~K., Minz, S., and Thakur, M.~P. (2019).
\newblock Rice plant disease classification using transfer learning of deep
  convolution neural network.
\newblock {\em International Archives of the Photogrammetry, Remote Sensing \&
  Spatial Information Sciences}.

\bibitem[Simonyan et~al., 2013]{simonyan2013deep}
Simonyan, K., Vedaldi, A., and Zisserman, A. (2013).
\newblock Deep inside convolutional networks: Visualising image classification
  models and saliency maps.
\newblock {\em arXiv preprint arXiv:1312.6034}.

\bibitem[Snoek et~al., 2012]{snoek2012practical}
Snoek, J., Larochelle, H., and Adams, R.~P. (2012).
\newblock Practical bayesian optimization of machine learning algorithms.
\newblock In {\em Advances in neural information processing systems}, pages
  2951--2959.

\bibitem[Springenberg et~al., 2014]{springenberg2014striving}
Springenberg, J.~T., Dosovitskiy, A., Brox, T., and Riedmiller, M. (2014).
\newblock Striving for simplicity: The all convolutional net.
\newblock {\em arXiv preprint arXiv:1412.6806}.

\bibitem[Szegedy et~al., 2016]{szegedy2016rethinking}
Szegedy, C., Vanhoucke, V., Ioffe, S., Shlens, J., and Wojna, Z. (2016).
\newblock Rethinking the inception architecture for computer vision.
\newblock In {\em Proceedings of the IEEE conference on computer vision and
  pattern recognition}, pages 2818--2826.

\bibitem[Wang et~al., 2019]{wang2019deep}
Wang, H., Won, D., and Yoon, S.~W. (2019).
\newblock A deep separable neural network for human tissue identification in
  three-dimensional optical coherence tomography images.
\newblock {\em IISE Transactions on Healthcare Systems Engineering},
  9(3):250--271.

\bibitem[Wu et~al., 2017]{wu2017characterization}
Wu, Y., Yu, L., Xiao, N., Dai, Z., Li, Y., Pan, C., Zhang, X., Liu, G., and Li,
  A. (2017).
\newblock Characterization and evaluation of rice blast resistance of chinese
  indica hybrid rice parental lines.
\newblock {\em The Crop Journal}, 5(6):509--517.

\bibitem[Zeiler and Fergus, 2014]{zeiler2014visualizing}
Zeiler, M.~D. and Fergus, R. (2014).
\newblock Visualizing and understanding convolutional networks.
\newblock In {\em European conference on computer vision}, pages 818--833.
  Springer.

\bibitem[Zhang et~al., 2018]{zhang2018medical}
Zhang, Q., Wang, H., Lu, H., Won, D., and Yoon, S.~W. (2018).
\newblock Medical image synthesis with generative adversarial networks for
  tissue recognition.
\newblock In {\em 2018 IEEE International Conference on Healthcare Informatics
  (ICHI)}, pages 199--207. IEEE.

\end{thebibliography}


\end{document}